\documentclass{article}

\PassOptionsToPackage{numbers, compress}{natbib}
\usepackage[final]{neurips_2025}

\usepackage[utf8]{inputenc}
\usepackage[T1]{fontenc}
\usepackage{hyperref}
\hypersetup{
    colorlinks=true,
    linkcolor=blue,
    citecolor=blue,      
    urlcolor=blue
}
\usepackage{url}
\usepackage{booktabs}
\usepackage{amsfonts}
\usepackage{nicefrac}
\usepackage{microtype}
\usepackage{xcolor}
\usepackage{bbding}
\usepackage{pifont}

\usepackage{multirow}
\usepackage{graphicx}
\usepackage{amsmath}
\usepackage{amsfonts}
\usepackage{amssymb}
\usepackage{amsthm}
\usepackage{bm}

\usepackage{xcolor}
\usepackage{array}
\usepackage[table]{xcolor}
\usepackage{colortbl}

\newtheorem{theorem}{Theorem}[section]
\newtheorem{corollary}[theorem]{Corollary}
\newtheorem{assumption}[theorem]{Assumption}

\usepackage{subcaption}

\usepackage{enumitem}

\usepackage{algorithm}
\usepackage{algpseudocode}

\usepackage{wrapfig}

\title{FlyLoRA: Boosting Task Decoupling and Parameter Efficiency via Implicit Rank-Wise Mixture-of-Experts} 

\author{
  Heming Zou\textsuperscript{1}\thanks{Equal contribution} \quad
  Yunliang Zang\textsuperscript{2}\footnotemark[1] \quad
  Wutong Xu\textsuperscript{1} \quad
  Yao Zhu\textsuperscript{1} \quad
  Xiangyang Ji\textsuperscript{1}\thanks{Corresponding author} \\
  \textsuperscript{1}Department of Automation, Tsinghua University \\
  \textsuperscript{2}Academy of Medical Engineering and Translational Medicine, Tianjin University \\
  \texttt{\{zouhm24, xwt22\}@mails.tsinghua.edu.cn} \\
  \texttt{yunliangzang@tju.edu.cn, ee\_zhuy@zju.edu.cn} \\
  \texttt{xyji@tsinghua.edu.cn} \\
}

\begin{document}

\maketitle

\begin{abstract}
    Low-Rank Adaptation (LoRA) is a widely used parameter-efficient fine-tuning method for foundation models, but it suffers from parameter interference, resulting in suboptimal performance. Although Mixture-of-Experts (MoE)-based LoRA variants show promise in mitigating intra-task correlations in single-task instruction tuning, they introduce additional router parameters and remain ineffective in multi-task model merging where inter-task interference arises. Inspired by the fly olfactory circuit, we propose FlyLoRA, an implicit MoE-based LoRA variant that introduces: (1) rank-wise expert activation in the up-projection matrix, and (2) an implicit router that unifies expert routing and down-projection, where a frozen sparse random projection matrix replaces the traditional dense trainable version. This design resolves the trade-off between intra-task decorrelation and computational efficiency by eliminating the need for an explicit router, while inherently mitigating inter-task interference due to the orthogonality property of random matrices. Extensive experiments across four domains---general knowledge understanding, scientific question answering, mathematical reasoning, and code generation---demonstrate consistent performance improvements over existing methods. Beyond empirical gains, FlyLoRA highlights how biological structures can inspire innovations in AI technologies. Code is available at \url{https://github.com/gfyddha/FlyLoRA}.
\end{abstract}

\section{Introduction}
Foundation models have demonstrated remarkable cross-domain capabilities with the scaling of model parameters \cite{achiam2023gpt, brown2020language, devlin2019bert, liu2024deepseek, ouyang2022training, radford2018improving, radford2019language}. To enhance their performance on downstream tasks, Supervised Fine-Tuning (SFT) has become a typical post-training approach. However, full-parameter fine-tuning (Full FT) incurs prohibitive computational overhead and storage costs, making customized deployment impractical for most individual users. To address this issue, Parameter-Efficient Fine-Tuning (PEFT) \cite{houlsby2019parameter, hu2022lora, lester2021power, li2021prefix, liu2022few, liu2024gpt, xiao2025dynaprompt, zaken2021bitfit} has emerged as a widely adopted technique that significantly reduces resource consumption by keeping pre-trained weights frozen while fine-tuning only a small set of additional injected parameters.

Low-Rank Adaptation (LoRA) \cite{hu2022lora} is one of the most prominent PEFT methods. By leveraging the intrinsic low-dimensional properties of large language models \cite{aghajanyan2020intrinsic, li2018measuring}, LoRA approximates the parameter matrix update $\Delta\bm{W}\in\mathbb{R}^{m\times n}$ as the product of two low-rank matrices, $\bm{B}\in\mathbb{R}^{m\times r}$ and $\bm{A}\in\mathbb{R}^{r\times n}$, where $r\ll\min(m,n)$. This method preserves much of the capability of Full FT across most tasks while substantially reducing both memory requirements and computational overhead.

However, to achieve strong performance on complex tasks, LoRA typically requires much higher ranks, which contradicts PEFT's core goal of efficiency \cite{jiang2024mora, liu2024dora}. Moreover, interference within LoRA's ranks can impair training \cite{tian2024hydralora}, leading to issues such as hallucination \cite{gekhman2024does} and gradient explosion \cite{pennington2017resurrecting}, thereby largely limiting its potential. We refer to this challenge as \textbf{intra-task interference}. Meanwhile, foundation models often need to integrate multiple capabilities to handle complex downstream tasks, but retraining on multi-domain corpora is expensive \cite{shen2024go4align}, particularly when several specialized models are already available. Consequently, model merging \cite{chronopoulou2023adaptersoup, huang2023lorahub, ilharco2022editing, matena2022merging} is widely used to combine LoRA components trained on different domains in a training-free manner. Arising from conflicts between different components, this process introduces another challenge: \textbf{inter-task interference}.

\begin{figure}[t]
    \centering
    \begin{subfigure}{0.32\textwidth}
        \centering
        \includegraphics[width=\textwidth]{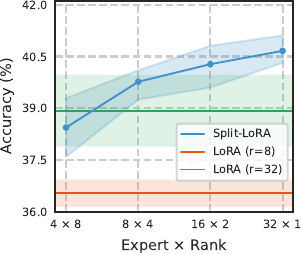}
        \caption{}
    \end{subfigure} %\hspace{1em}
    \begin{subfigure}{0.32\textwidth}
        \centering
        \includegraphics[width=\textwidth]{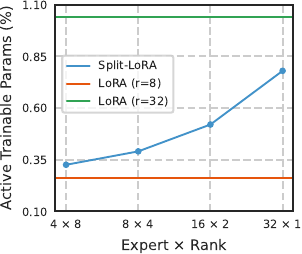}
        \caption{}
    \end{subfigure} %\hspace{1em}
    \begin{subfigure}{0.32\textwidth}
        \centering
        \includegraphics[width=\textwidth]{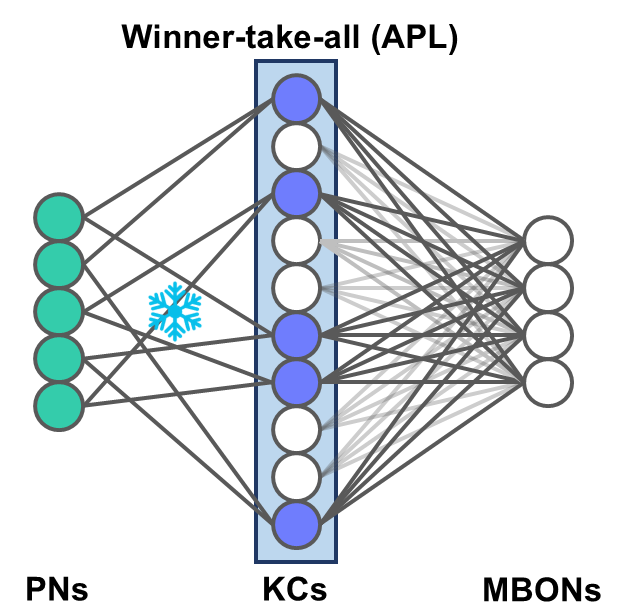}
        \caption{} 
    \end{subfigure}
    \caption{\textbf{(a)} Accuracy comparison under a fixed total rank $r=32$ and activation rank $k=8$. Finer-grained rank allocation (from $4$ experts $\times~8$ rank to $32$ experts $\times~1$ rank) yields consistent performance gains. \textbf{(b)} Activated trainable parameters (relative to Full FT) under the same budget. Increasing expert granularity leads to a monotonic rise in activated parameters due to router overhead. \textbf{(c)} Schematic of the fly olfactory circuit. Odor signals in projection neurons (PNs) are randomly projected to Kenyon cells (KCs), with each KC connecting to a fixed number of PNs (but not all), forming sparse connections. These signals are then selectively projected to mushroom body output neurons (MBONs), while lateral inhibition from an anterior paired lateral (APL) neuron suppresses weak KC-MBON connections, implementing a winner-take-all strategy. Thus, the number of activated KCs is much smaller than the total dimension of the KC layer.
    }
    \label{fig:fig1}
    \vspace{-10pt}
\end{figure}

To address intra-task interference, several studies have incorporated the Mixture-of-Experts (MoE) architecture \cite{fedus2022switch, jacobs1991adaptive, lepikhin2020gshard, shazeer2017outrageously, wang2022learning, wang2024separable} into LoRA \cite{dou2023loramoe, feng2024mixture, gao2024higher, li2024mixlora, muqeeth2023soft, tian2024hydralora, wu2024mixture, zadouri2023pushing}, where each expert learns specialized knowledge to partially achieve task decoupling. We refer to these approaches as \textbf{MoE-based LoRA methods}. They replace the original low-rank matrices with multiple experts and use a dynamic router to selectively activate them. By leveraging redundant parameters and sparse activations, these methods keep the computational budget comparable to LoRA with fewer total ranks. However, they may still suffer from interference within each expert. To investigate this issue, we conduct pilot studies on MMLU using Llama-3.1-8B, adopting Split-LoRA as a representative MoE-based method. As shown in Figure \ref{fig:fig1}(a), finer-grained rank allocation yields consistent performance improvements under a fixed budget. However, as illustrated in Figure \ref{fig:fig1}(b), pushing expert granularity to extremes also increases the number of activated trainable parameters due to additional router overhead. This trade-off makes it difficult to achieve both high performance and efficiency. Meanwhile, resolving inter-task interference has received limited emphasis in existing MoE-based LoRA methods.

Therefore, we seek to design an improved MoE-based LoRA variant that simultaneously achieves:

\hspace{1em} $\bullet$ \emph{~Reduced parameter interference among different ranks within a single LoRA component;}

\hspace{1em} $\bullet$ \emph{~Reduced parameter interference between different LoRA components;}

\hspace{1em} $\bullet$ \emph{~Reduced activated trainable parameters in routers.}

Inspired by the fly olfactory circuit \cite{caron2013random, dasgupta2017neural, li2025seemingly, lin2014sparse, stevens2015fly, zou2025structural}, which shows strong similarity to MoE-based LoRA, we introduce an implicit router to mitigate the trade-off between intra-task interference and efficiency. As shown in Figure \ref{fig:fig1}(c), this leads to the design of \textbf{FlyLoRA}, which (1) treats matrix $\bm{A}$ as a frozen sparse random projection that maps inputs into a higher-rank space (e.g., $r=32$ vs. $r=8$); and (2) simulates the bio-inspired ``winner-take-all'' mechanism by activating $k$ rank-1 experts in $\bm{B}$ linked to the top-$k$ magnitudes after projection by $\bm{A}$. This unifies the roles of $\bm{A}$ and router $\bm{G}$ into a single frozen projection, jointly performing down-projection and expert selection. Without explicit router parameters, the resulting implicit rank-wise MoE structure maintains computational efficiency while reducing intra-task interference. Moreover, we theoretically show that distinct random projections $\bm{A}_i$ and $\bm{A}_j$ from different LoRA components naturally map task updates into approximately orthogonal subspaces, thereby alleviating inter-task interference.

In summary, the core contributions of our proposed FlyLoRA framework are:
% \begin{enumerate}[topsep=0pt,itemsep=0ex,leftmargin=3ex]
%     \item \textbf{Efficient Intra-task Decoupling:} By using implicit rank-wise MoE, we enable finer expert allocation with reduced parameter interference in single-task scenarios. Additionally, FlyLoRA surpasses MoE-based LoRA in efficiency by eliminating the need for router parameters.
%     \item \textbf{Efficient Inter-task Decoupling:} In multi-task model merging scenarios, different random projections $\bm{A}_i$ and $\bm{A}_j$ naturally form approximately orthogonal subspaces. This inherent property ensures different LoRA components operate in uncorrelated subspaces, thus achieves decoupling.  
%     \item \textbf{Neuroscience-Inspired Design:} The efficacy of our algorithm, combined with its structural alignment with the fly olfactory circuit, establishes a promising bridge between neuroscience and artificial intelligence.
% \end{enumerate}

\hspace{1em} $\bullet$ \textbf{Efficient Intra-task Decoupling:} By using implicit rank-wise MoE, we enable finer expert allocation with reduced parameter interference in single-task scenarios. Additionally, FlyLoRA surpasses MoE-based LoRA in efficiency by eliminating the need for router parameters.

\hspace{1em} $\bullet$ \textbf{Efficient Inter-task Decoupling:} In multi-task model merging scenarios, different random projections $\bm{A}_i$ and $\bm{A}_j$ naturally form approximately orthogonal subspaces. This inherent property ensures different LoRA components operate in uncorrelated subspaces, thus achieves decoupling.  

\hspace{1em} $\bullet$ \textbf{Neuroscience-Inspired Design:} The efficacy of our algorithm, combined with its structural alignment with the fly olfactory circuit, establishes a promising bridge between neuroscience and artificial intelligence.

\section{Revisiting MoE-based LoRA Methods} \label{sec:revisit_moelora}

\subsection{Preliminaries}
LoRA (visualized in Figure \ref{fig:fig2}(a)) simulates weight updates during fine-tuning by decomposing the update matrix into two learnable low-rank matrices. Given a pretrained weight matrix $\bm{W}_0\in\mathbb{R}^{m\times n}$, the parameter update is computed as:
\begin{equation}
    \bm{W}'=\bm{W}_0+\Delta\bm{W}=\bm{W}_0+\frac{\alpha}{r}\bm{BA},
\end{equation}
where $\bm{B}\in\mathbb{R}^{m\times r}$, $\bm{A}\in\mathbb{R}^{r\times n}$, and the rank $r\ll \min(m, n)$. The scaling factor $\alpha$ is typically set to $2r$. For an input embedding $\bm{x}\in\mathbb{R}^n$, the forward pass becomes:
\begin{equation} \label{eq:lora}
    f_\text{LoRA}(\bm{x})=\bm{W}'\bm{x}=\bm{W}_0\bm{x}+\frac{\alpha}{r}\bm{BAx}.
\end{equation}
Here, $\bm{W}_0$ remains frozen during training, while only $\{\bm{A},\bm{B}\}$ are updated. This approach reduces the number of trainable parameters from $\mathcal{O}(mn)$ to $\mathcal{O}(r(m+n))$, thereby achieving higher parameter efficiency. The low-rank structure allows LoRA to maintain stable performance while significantly reducing both computational overhead and GPU memory requirements during fine-tuning.

\subsection{MoE-based LoRA Framework} \label{Sec:MoELoRA}
\begin{figure}[t]
    \centering
    \begin{subfigure}{0.32\textwidth}
        \centering
        \includegraphics[width=\textwidth]{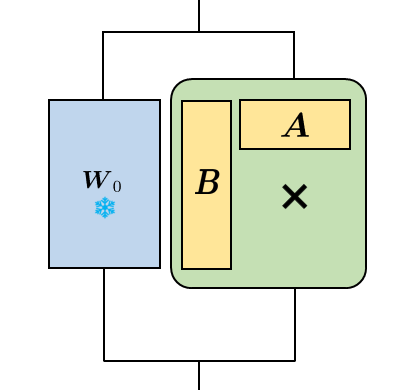}
        \caption{LoRA}
    \end{subfigure}
    \begin{subfigure}{0.32\textwidth}
        \centering
        \includegraphics[width=\textwidth]{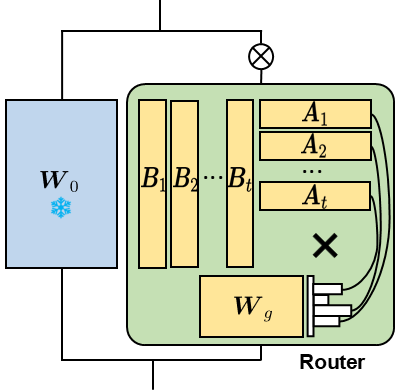}
        \caption{MoE-based LoRA}
    \end{subfigure}
    \begin{subfigure}{0.32\textwidth}
        \centering
        \includegraphics[width=\textwidth]{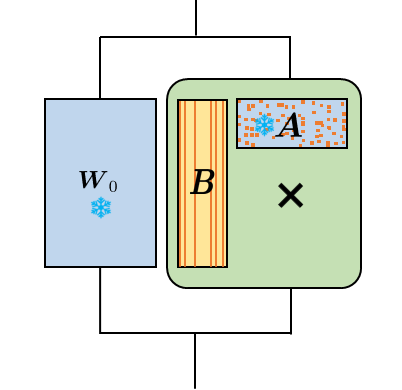}
        \caption{FlyLoRA}
    \end{subfigure}
    \caption{\textbf{Schematic illustrations of different LoRA variants.} \textbf{(a)} LoRA employs low-rank matrices $\bm{A}$ and $\bm{B}$ to simulate weight updates, where each row of $\bm{A}$ is fully connected to the corresponding column of $\bm{B}$. \textbf{(b)} MoE-based LoRA decomposes the updates into multiple small experts $\{\bm{A}_i, \bm{B}_i\}_{i=1}^N$ and uses a router to determine which experts should be activated. \textbf{(c)} FlyLoRA unifies the down-projection and router into a frozen matrix $\bm{A}$ and selectively activates only the ranks in $\bm{B}$ linked to the top-$k$ magnitude activations after projection through $\bm{A}$.}
    \label{fig:fig2}
    \vspace{-10pt}
\end{figure}

The MoE paradigm (visualized in Figure \ref{fig:fig2}(b)) extends LoRA by decomposing the low-rank adaptation into $N$ specialized experts. Each expert $\bm{E}_i$ is parameterized by a pair of matrices $\{\bm{B}_i\in\mathbb{R}^{m\times r_i}, \bm{A}_i\in\mathbb{R}^{r_i\times n}\}$, where $r_i$ denotes the expert-specific rank. The forward pass incorporates a gating mechanism $\bm{G}(\bm{x}):\mathbb{R}^n \rightarrow \mathbb{R}^N$ that dynamically routes inputs to activate the most relevant experts. Formally, the output combines the frozen pretrained weights $\bm{W}_0$ with a sparse combination of expert contributions:
\begin{equation} \label{eq:moe-lora}
    f_\text{MoE-LoRA}(\bm{x})=\bm{W}_0\bm{x}+\frac{\alpha}{r}\sum_{i=1}^N\bm{G}(\bm{x})_i\cdot\underbrace{\bm{B}_i\bm{A}_i\bm{x}}_{\bm{E}_i(\bm{x})},
\end{equation}
where the router $\bm{G}(\bm{x})$ typically follows a top-$k$ selection policy via a trainable projection $\bm{W}_g\in\mathbb{R}^{N\times n}$. For simplicity, we omit the activation function, formulating the router as:
\begin{equation} \label{eq:topk-router}
    \bm{G}(\bm{x})=\text{top-}k(\bm{W}_g\bm{x}).
\end{equation}
By activating only $k$ experts per input, this design maintains computational efficiency. The sparse routing strategy enables conditional computation, which expands the model's representational capacity without incurring a proportional increase in computational cost. In our work, we implement Split-LoRA under this framework as a minimal yet representative instantiation of MoE-based LoRA. Further implementation details are provided in Appendix \ref{App:Split-LoRA}.

\subsection{Pushing MoE-based LoRA Architecture to the Extreme} \label{Sec:moe_limit}
Comparing Eq.~\ref{eq:lora} and Eq.~\ref{eq:moe-lora} reveals that MoE-based LoRA can be viewed as a finer-grained, sparsely activated variant of LoRA, where the separation of experts mitigates task conflicts. Taking this decomposition to the extreme motivates our \textbf{rank-wise expert} design, where each expert governs a single rank, achieving the best decorrelating effect (see Figure \ref{fig:fig1}(a)). Formally, for a rank-$r$ LoRA, the matrices $\bm{A}$ and $\bm{B}$ can be decomposed into $r$ rank-$1$ components:
\begin{equation} \label{eq:rankwise-moe-lora}
    f_\text{rank-wise-LoRA}(\bm{x})=\bm{W}_0\bm{x}+\frac{\alpha}{r}\sum_{i=1}^r\bm{G}(\bm{x})_i\cdot\underbrace{\bm{b}_i\bm{a}_i\bm{x}}_{\bm{E}_i(\bm{x})},
\end{equation}
with $\bm{a}_i=\bm{A}[i,:]\in\mathbb{R}^{1\times n}$ and $\bm{b}_i=\bm{B}[:,i]\in\mathbb{R}^{m\times 1}$.

However, this approach introduces a scalability challenge: the router's linear layer $\bm{W}_g\in\mathbb{R}^{N\times n}$ grows linearly with the number of experts $N$ (see Figure \ref{fig:fig1}(b)). Under a fixed total rank, finer-grained experts with larger $N$ make the explicit routing mechanism computationally prohibitive, undermining the efficiency gains of sparse activation.

To overcome this limitation, we seek an \textbf{implicit routing mechanism} that eliminates the need for the explicit router parameter $\bm{W}_g$ entirely. This entails finding a proxy that leverages intrinsic signals within the model to select the top-$k$ experts, effectively approximating the function of the original router $\bm{G}$. To address this, we draw inspiration from the perspective of Singular Value Decomposition (SVD), which can also be viewed as a rank-wise decomposition. In SVD, the low-rank update matrix $\Delta \bm{W}$ can be decomposed as $\Delta \bm{W} = \sum_{i=1}^r \sigma_i \bm{u}_i \bm{v}_i^\top$, where $\sigma_i$ denotes the $i$-th singular value (indicating the importance of the corresponding component), $\bm{u}_i$ is the $i$-th left-singular vector, and $\bm{v}_i$ is the $i$-th right-singular vector. Each component $\sigma_i \bm{u}_i \bm{v}_i^\top$ is a rank-1 update. The Eckart-Young-Mirsky theorem \cite{eckart1936approximation} guarantees that the top-$k$ components, selected based on the magnitude of $\sigma_i$, provide the best rank-$k$ approximation to the original rank-$r$ matrix in terms of Frobenius norm, thereby capturing the most salient features with minimal reconstruction error. While exact SVD is computationally prohibitive and thus impractical in our framework, this insight naturally suggests that the magnitude of each rank-$1$ term, $\|\bm{b}_i\bm{a}_i\bm{x}\|$ in Eq.~\ref{eq:rankwise-add-lora}, approximately reflects its importance:
\begin{equation} \label{eq:rankwise-add-lora}
    f_\text{LoRA}(\bm{x})=\bm{W}_0\bm{x}+\frac{\alpha}{r}\sum_{i=1}^r\bm{b}_i\bm{a}_i\bm{x}.
\end{equation}
Nevertheless, a naive approach of first computing all $r$ terms $\bm{b}_i\bm{a}_i\bm{x}$ and then selecting the top-$k$ would also forfeit the computational benefits of sparse activation, as the cost of computing all terms remains $\mathcal{O}(rmn)$. This necessitates a routing strategy that can identify the most important experts \emph{before} fully computing their outputs. Furthermore, beyond efficient routing, another critical limitation of existing MoE-based LoRA methods is their lack of inherent support for multi-task deployment. When merging models already fine-tuned on different tasks, interference between LoRA adapters often leads to significant performance degradation, as the underlying architecture does not structurally encourage task-specific updates to reside in orthogonal or non-overlapping parameter subspaces.

These dual challenges motivate the following two key design requirements for an improved MoE-based LoRA framework:

\hspace{1em} $\bullet$ \emph{Implicit magnitude-based router for top-$k$ activation, without explicit router parameters, enabling expert selection prior to full computation;}

\hspace{1em} $\bullet$ \emph{Native support for training-free model merging through architectural properties that promote inter-task interference mitigation.}

\section{FlyLoRA} \label{Sec:flylora}
Inspired by the fly olfactory circuit (Figure \ref{fig:fig1}(c)), whose neural architecture inherently meets our requirements for MoE-based LoRA variants, we propose FlyLoRA (visualized in Figure \ref{fig:fig2}(c)). Section \ref{sec:formulation} presents its formal design, while subsequent sections analyze its key advantages: Section \ref{sec:distance_preserving} shows how a fixed $\bm{A}$ acts as an implicit router, Section \ref{sec:gradient_decoupling} demonstrates intra-task decoupling, and Section \ref{sec:inter_task_orthgonal} establishes inherent support for inter-task decoupling in model merging.

\subsection{Formulation of FlyLoRA} \label{sec:formulation}
In FlyLoRA, the matrix $\bm{A}\in\mathbb{R}^{r\times n}$ is sparse and frozen. It is randomly initialized at the beginning and remains frozen during training, implementing an intrinsic top-$k$ operation in the projection space $\mathbb{R}^r$ for implicit routing. Given an input token $\bm{x}\in\mathbb{R}^n$, this process is formulated as:
\begin{equation}\label{Eq:projection}
    \bm{y}'=\text{top-}k(\bm{y}) = \text{top-}k\left(\bm{A}\bm{x}\right),
\end{equation}
where each row of $\bm{A}$ contains exactly $p$ ($p<n$) non-zero entries independently sampled from $\mathcal{N}(0, \frac{1}{r^2})$ (a widely used standard initialization). We define the sparsity ratio as $\rho=\frac{p}{n}$. After projection through $\bm{A}$, only the columns $\bm{b}_i\in\mathbb{R}^m$ ($i\in\{1,\ldots, r\}$) in the up-projection matrix $\bm{B}\in\mathbb{R}^{m\times r}$ linked to dimensions with top-$k$ ($k<r$) magnitudes in $\bm{Ax}\in\mathbb{R}^r$ are activated. Formally:
\begin{equation}
    [\bm{By}']_i= \begin{cases}
    [\bm{By}]_i & \text{if the magnitude of } [\bm{y}]_i \text{ is among the top-$k$ values of } \bm{y}, \\
    0 & \text{otherwise.}
    \end{cases}
\end{equation}
To enhance training stability in this MoE structure, we incorporate a simple expert-wise bias term $\bm{d}\in\mathbb{R}^r$ for loss-free load balancing, following \cite{liu2024deepseek}. This auxiliary term is updated manually via:
\begin{equation} \label{eq:load_balancing}
    \bm{d}_i\leftarrow \bm{d}_i+u\cdot \text{sign}(\bar{\bm{c}_i}-\bm{c}_i),
\end{equation} 
where $u$ is a small learning rate, $\bar{\bm{c}_i}$ represents the expected assignment frequency for expert $i$, $\bm{c}_i$ tracks the actual assignment count, and $\text{sign}(\cdot)$ denotes the sign function. This bias term $\bm{d}$ is added to $\bm{Ax}$ in expert selection to promote the activation of under-activated experts and suppress over-activated experts, thereby achieving load balancing. Thus, the activated experts are selected by:
\begin{equation} \label{eq:topk}
    \mathcal{I}_{\text{top}k}=\big\{i_1,\dots,i_k\big\}\quad\text{where}\quad i_j=\mathop{\arg\max}\limits_{i\notin\{i_1,\dots,i_{j-1}\}}\big(\bm{Ax}+\bm{d}\big)_i.
\end{equation}
The forward pass is then computed as:
\begin{equation} \label{eq:flylora}
    f_\text{FlyLoRA}(\bm{x})=\bm{W}_0\bm{x}+\Delta\bm{Wx}=\bm{W}_0\bm{x}+\frac{\alpha}{r}\sum_{i=1}^{r}\mathbb{I}(i\in\mathcal{I}_{\text{top}k})\cdot\bm{b}_i\bm{a}_i\bm{x},
\end{equation}
where $\mathbb{I}(\cdot)$ denotes the indicator function.

\subsection{Fixed Sparse Random Projection as Implicit Router} \label{sec:distance_preserving}

\begin{figure}[t]
    \centering
    \begin{subfigure}{0.32\textwidth}
        \centering
        \includegraphics[width=\textwidth]{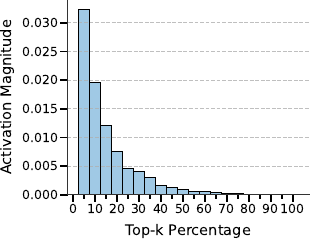}
        \caption{Activation distribution}
    \end{subfigure}
    \begin{subfigure}{0.32\textwidth}
        \centering
        \includegraphics[width=\textwidth]{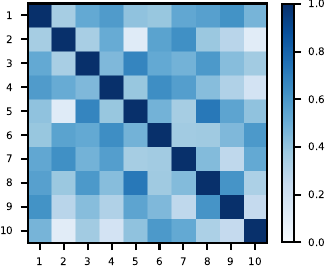}
        \caption{LoRA-FA gradient correlation}
    \end{subfigure}
    \begin{subfigure}{0.32\textwidth}
        \centering
        \includegraphics[width=\textwidth]{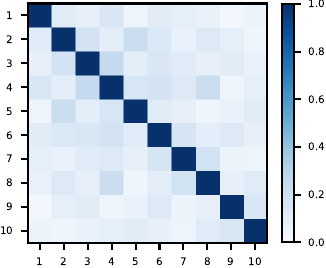}
        \caption{FlyLoRA gradient correlation}
    \end{subfigure}
    \caption{\textbf{(a)} Activation value magnitude distribution across dimensions, showing the mean activation strength at different top-$k$ selection percentages. \textbf{(b-c)} Gradient correlation matrices of (b) LoRA-FA$_{(r=32)}$ versus (c) FlyLoRA$_{(k=8)}$'s $\bm{B}$ matrices ($10$ randomly sampled columns). For a simplified illustration, we use the LoRA module of q\_proj in the middle layer of Llama-3.1-8B on MMLU.}
    \label{fig:fig3}
    \vspace{-10pt}
\end{figure}
The core objective of the MoE router is to select $k$ out of $r$ experts that best approximate the effect of using all experts. However, as established in Sec. \ref{Sec:moe_limit}, it is computationally impractical to determine the selection based on $\|\bm{b}_i\bm{a}_i\bm{x}\|$. We therefore seek to perform the selection using only $\bm{a}_i\bm{x}\in\mathbb{R}$ as a surrogate. In FlyLoRA, since we fix $\bm{A}$ as a sparse random projection, computing $\bm{a}_i\bm{x}$ is as efficient as in standard LoRA$_{(r=k)}$. We theoretically prove that this projection preserves pairwise distances (see Theorem \ref{thm:distance_preserving}), and thus can serve as an effective implicit router.
\begin{theorem} \label{thm:distance_preserving}
    Given the matrix $\bm{A}\in\mathbb{R}^{r\times n}$ with each row having exactly $p$ non-zero entries randomly sampled from $\mathcal{N}(0, \frac{1}{r^2})$, for any $\epsilon>0$,
    $$\mathbb{P}\left( (1-\epsilon) \|\bm{x}-\bm{y}\|^2 \leq \frac{1}{r\sigma^2}\|\bm{A}\bm{x}- \bm{A}\bm{y}\|^2 \leq (1+ \epsilon)  \|\bm{x}-\bm{y}\|^2\right) \geq 1-e^{-(\epsilon^2 - \epsilon^3) \frac{r}{4}} - e^{- \frac{(\epsilon^2 - \epsilon^3)r}{2(\frac{3p}{n} + 1)}},$$
    for any input embeddings $\bm{x},\bm{y}\in\mathbb{R}^n$, where $\sigma^2=\frac{p}{nr^2}$. A detailed proof is provided in Appendix \ref{App:distance_preserving}.
\end{theorem}
This theorem establishes a probabilistic guarantee for the approximate preservation of Euclidean distances under a sparse random projection. Notably, the concentration bound tightens with reduced sparsity ($p/n$) and larger rank ($r$), which aligns with intuitive expectation. We further present its robustness empirically via hyperparameter sensitivity analysis in Section \ref{sec:ablation_sensitivity}.

Building on this property, we posit that FlyLoRA can self-select experts based on the values of $\bm{a}_i\bm{x}$, which aligns with the finding in \cite{lv2025autonomy} that ``an expert is aware of its own capacity to effectively process a token, an awareness reflected in the scale of its internal activations.'' Unlike traditional trainable MoE routers that explicitly learn routing weights, FlyLoRA leverages the fixed geometry of $\bm{A}$ to perform implicit, activation-driven routing. This design not only eliminates the difficulty of learning routing parameters and avoids the separation between the router's decision-making and the experts' execution, but also reduces training instability by removing the stochasticity in router optimization. Because the projection $\bm{A}$ preserves pairwise distances, two semantically similar inputs $\bm{x}_i$ and $\bm{x}_j$ are mapped to nearby low-dimensional representations $\bm{A}\bm{x}_i$ and $\bm{A}\bm{x}_j$, and therefore routed to similar experts, while dissimilar inputs are routed to different experts. This geometry-induced consistency helps mitigate expert representation homogenization in MoE models, enabling each expert to focus on specialized knowledge, reduce internal conflicts, and improve sample efficiency (as supported by previous studies, e.g., \cite{chowdhury2023patch}). In this sense, FlyLoRA resembles a variant of the hash router \cite{roller2021hash}, which also achieves lightweight and stable expert assignment through a fixed mapping. Consequently, the top-$k$ operation naturally selects the most important experts according to the magnitudes of $\bm{a}_i\bm{x}$. In Figure \ref{fig:fig3}(a), empirically around top-25\% of dimensions account for more than 80\% ``energy''. Thus, it typically does not cause a large performance drop.

\subsection{Gradient Decoupling via Top-k Sparsity} \label{sec:gradient_decoupling}
In the FlyLoRA framework, only matrix $\bm{B}$ requires updating. We theoretically demonstrate that our rank-wise expert allocation strategy, induced by top-$k$ selection, inherently reduces gradient covariance between distinct experts, thus mitigating intra-task interference. Our analysis begins with Assumption \ref{asm:Uniform_Sparse_Activation} describing the sparsity pattern of activations. For analytical convenience, we consider a simplified condition where the top-$k$ operation randomly selects $k$ out of $r$ columns for activation.
\begin{assumption}[Uniform Sparse Activation] \label{asm:Uniform_Sparse_Activation}
    During the top-$k$ operation, each training sample activates exactly $k$ columns of the parameter matrix $\bm{B}\in\mathbb{R}^{m \times r}$, with uniform selection probability $p=\frac{k}{r}$ per column.
\end{assumption}
Based on Assumption \ref{asm:Uniform_Sparse_Activation}, we derive Theorem \ref{thm:covariance_reduction} (proof in Appendix \ref{App:topk_decoupling}).
\begin{theorem}[Covariance Reduction Under top-$k$] \label{thm:covariance_reduction}
    Let $\tilde{\bm{\Sigma}}$ and $\bm{\Sigma}$ denote the gradient covariance matrices with and without top-$k$ activation. When $r > k$, off-diagonal entries scale as:
    $$
    \mathbb{E}[\tilde{\bm{\Sigma}}_{(i,j)}] \approx \mathbb{E}[\bm{\Sigma}_{(i,j)}] \cdot \frac{k^2}{r^2}, \quad \forall i \neq j.
    $$
\end{theorem}
This $\mathcal{O}(k^2/r^2)$ reduction factor quantifies how top-$k$ sparsity promotes parameter decoupling by suppressing interference terms. When $k=1$ (only one rank is activated), the off-diagonal covariance almost vanishes, achieving full decoupling; when $k=r$ (all ranks are activated, degenerating to LoRA-FA \cite{zhang2023lora}), it recovers the dense training regime. Theorem \ref{thm:covariance_reduction} is proved in Appendix \ref{App:topk_decoupling}. To empirically validate this theoretical result, we visualize the gradient correlation patterns of LoRA-FA and FlyLoRA in Figure \ref{fig:fig3}(b) and (c), where correlations are computed using 10 randomly selected gradient columns (see heatmap visualization). The observed sparsity pattern strongly supports our theoretical prediction of reduced off-diagonal covariance under top-$k$ selection.

\subsection{Inter-Task Orthogonality in Model Merging} \label{sec:inter_task_orthgonal}
Traditional LoRA model merging often suffers from parameter interference when combining task-specific components through weight averaging:
\begin{equation} \label{eq:lora_merging}
    \bm{W}'=\bm{W}_0 + \sum_{i=1}^t w_i\bm{B}_i\bm{A}_i.
\end{equation}
We analyze how FlyLoRA's inherent subspace orthogonality enables effective multi-task model merging. We derive Theorem \ref{thm:subspace_orthogonality} (proof in Appendix \ref{App:subspace_orthogonality}).
\begin{theorem}[Approximate Subspace Orthogonality] \label{thm:subspace_orthogonality}
    For independent random matrices $\bm{A}_i,\bm{A}_j\in \mathbb{R}^{r\times n}$ with sparse Gaussian entries ($\mathcal{N}(0, \frac{1}{r^2})$ for $p<n$ randomly selected entries per row), the following holds,
    \begin{enumerate}
    \item\textbf{Exact mean orthogonality}: $\mathbb{E}[\bm{A}_i \bm{A}_j^\top]=\bm{0}_{r\times r}$
    \item\textbf{Polynomially decaying correlations}: $\mathbb{P}(\| \bm{A}_i\bm{A}_j^\top\|_2\ge \epsilon r)\le \frac{p^2}{nr^2\epsilon^2}$
    \end{enumerate}
\end{theorem}
This theorem establishes that sparse random projections naturally induce nearly orthogonal subspaces. The residual correlation bound of order $\mathcal{O}(\frac{p^2}{nr^2})$ indicates that interference becomes negligible under practical parameter scales. This property directly leads to Corollary \ref{cor:merge} (proof in Appendix \ref{App:subspace_orthogonality}).
\begin{corollary} \label{cor:merge}
    Let $\bm{A}_i,\bm{A}_j \in \mathbb{R}^{r \times n}$ be fixed sparse random projections after initialization. Then for any learned matrices $\bm{B}_i\bm{A}_i$ and $\bm{B}_j\bm{A}_j$, they satisfy the pairwise orthogonality property:
    $$
    \langle \bm{B}_i\bm{A}_i, \bm{B}_j\bm{A}_j \rangle_F \approx 0 \quad \text{for} \quad i \neq j.
    $$
\end{corollary}
This orthogonal decomposition provides a key advantage for model merging: task-specific updates $\bm{B}_i\bm{A}_i$ occupy nearly orthogonal subspaces, thereby preventing destructive interference. The ``Pairwise Orthogonality'' property captures FlyLoRA's behavior during multi-task aggregation. According to geometric intuition that orthogonality facilitates model merging~\cite{ilharco2022editing, ortiz2023task, tang2023parameter}, and consistent with theoretical analyses in~\cite{li2025task, zeng2025efficient}, FlyLoRA's fixed sparse projection design aligns with this principle. Empirically (see Section~\ref{sec:multi-task}), the random projection $\bm{A}$ enables FlyLoRA to preserve task-specific performance after merging, whereas the learnable $\bm{A}$ in conventional LoRA exhibits significantly higher interference. A similar analysis of LoRA merging was conducted in a concurrent study \cite{zhang2025lori}.
% A concurrent work \cite{zhang2025lori} also follows this established principle.

\section{Experiments} \label{Sec:experiment}

\subsection{Experimental Setup}

\paragraph{Datasets and Backbones:}
We evaluate FlyLoRA's performance across four key domains: (1) \textit{general knowledge understanding} using the MMLU \cite{hendryckstest2021} benchmark with auxiliary training datasets for fine-tuning and test set for evaluation, (2) \textit{scientific question answering} using the ScienceQA \cite{lu2022learn} dataset for fine-tuning and evaluation, (3) \textit{mathematical reasoning} on GSM8K \cite{cobbe2021gsm8k} problems for fine-tuning and evaluation, and (4) \textit{code generation} assessed via CodeAlpaca-20k \cite{codealpaca} for training and HumanEval \cite{chen2021evaluating} for evaluation. Except for HumanEval, which is evaluated via pass@k metrics, others are evaluated via accuracy. All benchmarks are evaluated in a zero-shot manner. We examine our framework in both single-task configurations, training with these four datasets individually, and multi-task settings, where the LoRA components trained in single-task setup for each dataset are merged together in a training-free manner. Most experiments are conducted using Llama-3.1-8B \cite{grattafiori2024llama} and Qwen-2.5-7B \cite{yang2024qwen2}, respectively. See Appendix \ref{app:experimental_setting} for implementation details.

\paragraph{Baselines:}
For the single-task setup, we compare FlyLoRA against: (1) vanilla LoRA with identical activation ranks (LoRA$_{(r=8)}$) and total ranks (LoRA$_{(r=32)}$), and (2) representative MoE-based LoRA variants Split-LoRA$_{(4\times8)}$ (abbreviated for $4$ expert$\times8$ rank). For the multi-task setup, we benchmark FlyLoRA against them using weight averaging fusion, and several advanced merging techniques. Across all datasets, FlyLoRA$_{(k=8)}$ uses total rank $r=32$ but activates only $k=8$ ranks after the top-$k$ operation between $\bm{A}$ and $\bm{B}$, with the fixed sparse random $\bm{A}$'s sparsity ratio $\rho$ set to $8/32$.
\begin{table}[t]
    \centering
    \caption{
    \textbf{Performance Comparison of LoRA Variants in Single-task Evaluation.} We evaluate various methods across four benchmarks: MMLU, ScienceQA, GSM8K (accuracy), and HumanEval (Pass@k), with all metrics reported in percentage (\%). Param(\%) indicates the percentage of activated trainable parameters relative to Full FT. The best results are highlighted in \textbf{bold}.}
    \vspace{+2pt}
    \renewcommand\arraystretch{1.0}
    \resizebox{1.0\textwidth}{!}{
    \setlength{\tabcolsep}{1mm}{
    \begin{tabular}{ccccccccc}
        \toprule
        \multirow{2}{*}{\textbf{Model}} &\multirow{2}{*}{\textbf{Method}} &\multirow{2}{*}{\textbf{Param(\%)}} &\multirow{2}{*}{\textbf{MMLU}} &\multirow{2}{*}{\textbf{ScienceQA}} &\multirow{2}{*}{\textbf{GSM8K}} &\multicolumn{3}{c}{\textbf{HumanEval}} \\
        & & & & & &\textbf{Pass@1} &\textbf{Pass@5} &\textbf{Pass@10} \\
        \midrule
        \multirow{4}{*}{Llama-3.1-8B}
            &LoRA$_{(r=8)}$ &0.26 &36.53\scriptsize{$\pm$0.40} &91.39\scriptsize{$\pm$0.55} &55.34\scriptsize{$\pm$0.24} &29.13\scriptsize{$\pm$0.56} &52.28\scriptsize{$\pm$1.24} &61.67\scriptsize{$\pm$0.61} \\
            &LoRA$_{(r=32)}$ &1.03 &38.93\scriptsize{$\pm$1.04} &94.01\scriptsize{$\pm$0.17} &56.25\scriptsize{$\pm$0.29} &30.37\scriptsize{$\pm$1.06} &54.37\scriptsize{$\pm$0.39} &64.02\scriptsize{$\pm$0.94} \\
            &Split-LoRA$_{(4\times 8)}$ &0.33 &38.44\scriptsize{$\pm$0.69} &92.41\scriptsize{$\pm$0.54} &55.65\scriptsize{$\pm$0.47} &31.28\scriptsize{$\pm$1.52} &54.16\scriptsize{$\pm$1.12} &63.94\scriptsize{$\pm$0.89} \\
            \cmidrule{2-9}
            &\cellcolor{gray!20}\textbf{FlyLoRA$_{(k=8)}$} &\cellcolor{gray!20}\textbf{0.13} &\cellcolor{gray!20}\textbf{40.88\scriptsize{$\pm$1.61}} &\cellcolor{gray!20}\textbf{94.15\scriptsize{$\pm$0.36}} &\cellcolor{gray!20}\textbf{58.76\scriptsize{$\pm$0.74}} &\cellcolor{gray!20}\textbf{36.88\scriptsize{$\pm$1.91}} &\cellcolor{gray!20}\textbf{62.40\scriptsize{$\pm$1.82}} &\cellcolor{gray!20}\textbf{73.34\scriptsize{$\pm$1.24}} \\
        \midrule
        \multirow{4}{*}{Qwen-2.5-7B}
            &LoRA$_{(r=8)}$ &0.26 &49.84\scriptsize{$\pm$0.56} &92.84\scriptsize{$\pm$0.13} &77.01\scriptsize{$\pm$0.32} &47.20\scriptsize{$\pm$1.54} &78.89\scriptsize{$\pm$0.36} &85.94\scriptsize{$\pm$0.64} \\
            &LoRA$_{(r=32)}$ &1.05 &52.07\scriptsize{$\pm$0.31} &95.01\scriptsize{$\pm$0.21} &79.23\scriptsize{$\pm$0.22} &52.87\scriptsize{$\pm$1.79} &81.67\scriptsize{$\pm$1.14} &87.80\scriptsize{$\pm$0.72} \\
            &Split-LoRA$_{(4\times 8)}$ &0.33 &50.68\scriptsize{$\pm$1.06} &93.08\scriptsize{$\pm$0.41} &77.12\scriptsize{$\pm$0.76} &48.65\scriptsize{$\pm$1.18} &79.30\scriptsize{$\pm$0.91} &86.05\scriptsize{$\pm$0.44} \\
            \cmidrule{2-9}
            &\cellcolor{gray!20}\textbf{FlyLoRA$_{(k=8)}$} &\cellcolor{gray!20}\textbf{0.13} &\cellcolor{gray!20}\textbf{53.68\scriptsize{$\pm$0.47}} &\cellcolor{gray!20}\textbf{95.55\scriptsize{$\pm$0.18}} &\cellcolor{gray!20}\textbf{80.82\scriptsize{$\pm$0.56}} &\cellcolor{gray!20}\textbf{54.34\scriptsize{$\pm$2.13}} &\cellcolor{gray!20}\textbf{82.85\scriptsize{$\pm$0.52}} &\cellcolor{gray!20}\textbf{89.63\scriptsize{$\pm$0.55}} \\
        \bottomrule
    \end{tabular}}}
    \label{tab:single_results}
    \vspace{-5pt}
\end{table}
\begin{table}[t]
    \centering
    \caption{
    \textbf{Multi-task Performance Comparison Before and After Parameter Merging.} We evaluate LoRA variants across MMLU, ScienceQA, GSM8K (accuracy), and HumanEval (Pass@k) benchmarks. The table shows performance before merging, after merging, and the relative performance drop ($\Delta$\%). The best results are highlighted in \textbf{bold}.
    }
    \vspace{+2pt}
    \renewcommand\arraystretch{1.0}
    \resizebox{1.0\textwidth}{!}{
    \setlength{\tabcolsep}{1mm}{
    \begin{tabular}{ccccccccc}
        \toprule
        \multirow{2}{*}{\textbf{Model}} & 
        \multirow{2}{*}{\textbf{Method}} & 
        \multirow{2}{*}{\textbf{Merge Status}} & 
        \multirow{2}{*}{\textbf{MMLU}} & 
        \multirow{2}{*}{\textbf{ScienceQA}} & 
        \multirow{2}{*}{\textbf{GSM8K}} & 
        \multicolumn{3}{c}{\textbf{HumanEval}} \\
        & & & & & & \textbf{Pass@1} & \textbf{Pass@5} & \textbf{Pass@10} \\
        \midrule
        \multirow{13}{*}{Llama-3.1-8B} 
            & \multirow{3}{*}{LoRA$_{(r=8)}$} 
            & Before &36.53\scriptsize{$\pm$0.40} &91.39\scriptsize{$\pm$0.55} &55.34\scriptsize{$\pm$0.24} &29.13\scriptsize{$\pm$0.56} &52.28\scriptsize{$\pm$1.24} &61.67\scriptsize{$\pm$0.61} \\
            & & After &30.05\scriptsize{$\pm$0.82} &31.05\scriptsize{$\pm$2.38} &25.19\scriptsize{$\pm$2.36} &16.09\scriptsize{$\pm$3.15} &45.38\scriptsize{$\pm$1.62} &56.49\scriptsize{$\pm$2.13} \\
            & & $\Delta$ (\%) &-6.48 &-60.34 &-30.15 &-13.04 &-6.90 &-5.18 \\
            \cmidrule{2-9}
            & \multirow{3}{*}{LoRA$_{(r=32)}$} 
            & Before &38.93\scriptsize{$\pm$1.04} &94.01\scriptsize{$\pm$0.17} &56.25\scriptsize{$\pm$0.29} &30.37\scriptsize{$\pm$1.06} &54.37\scriptsize{$\pm$0.39} &64.02\scriptsize{$\pm$0.94} \\
            & & After &34.02\scriptsize{$\pm$1.32} &34.35\scriptsize{$\pm$1.42} &24.77\scriptsize{$\pm$0.94} &18.94\scriptsize{$\pm$1.48} &46.39\scriptsize{$\pm$1.74} &59.27\scriptsize{$\pm$1.19} \\
            & & $\Delta$ (\%) &-4.91 &-59.66 &-31.48 &-11.43 &-7.98 &-4.75 \\
            \cmidrule{2-9}
            & \multirow{3}{*}{Split-LoRA$_{(4\times8)}$} 
            & Before &38.44\scriptsize{$\pm$0.69} &92.41\scriptsize{$\pm$0.54} &55.65\scriptsize{$\pm$0.47} &31.28\scriptsize{$\pm$1.52} &54.16\scriptsize{$\pm$1.12} &63.94\scriptsize{$\pm$0.89} \\
            & & After &33.58\scriptsize{$\pm$1.16} &37.67\scriptsize{$\pm$1.06} &27.35\scriptsize{$\pm$1.10} &21.36\scriptsize{$\pm$1.08} &46.01\scriptsize{$\pm$0.87} &59.52\scriptsize{$\pm$0.95} \\
            & & $\Delta$ (\%) &-4.86 &-54.74 &-28.30 &-9.92 &-8.15 &-4.42 \\
            \cmidrule{2-9}
            & \multirow{3}{*}{FlyLoRA$_{(k=8)}$} 
            & Before &40.88\scriptsize{$\pm$1.61} &94.15\scriptsize{$\pm$0.36} &58.76\scriptsize{$\pm$0.74} &36.88\scriptsize{$\pm$1.91} &62.40\scriptsize{$\pm$1.82} &73.34\scriptsize{$\pm$1.24} \\
            & & After &38.86\scriptsize{$\pm$1.46} &51.10\scriptsize{$\pm$0.71} &36.95\scriptsize{$\pm$1.37} &32.61\scriptsize{$\pm$1.45} &56.59\scriptsize{$\pm$2.66} &69.76\scriptsize{$\pm$0.75} \\
            & &\cellcolor{gray!20}\textbf{$\Delta$ (\%)} &\cellcolor{gray!20}\textbf{-2.02} &\cellcolor{gray!20}\textbf{-43.05} &\cellcolor{gray!20}\textbf{-21.81} &\cellcolor{gray!20}\textbf{-4.27} &\cellcolor{gray!20}\textbf{-5.81} &\cellcolor{gray!20}\textbf{-3.58} \\
        \midrule
        \multirow{13}{*}{Qwen-2.5-7B} 
            & \multirow{3}{*}{LoRA$_{(r=8)}$} 
            & Before &49.84\scriptsize{$\pm$0.56} &92.84\scriptsize{$\pm$0.13} &77.01\scriptsize{$\pm$0.32} &47.20\scriptsize{$\pm$1.54} &78.89\scriptsize{$\pm$0.36} &85.94\scriptsize{$\pm$0.64} \\
            & & After &44.62\scriptsize{$\pm$1.23} &60.07\scriptsize{$\pm$2.18} &81.56\scriptsize{$\pm$1.48} &22.09\scriptsize{$\pm$2.68} &68.38\scriptsize{$\pm$0.86} &80.49\scriptsize{$\pm$0.77} \\
            & & $\Delta$ (\%) &-5.22 &-32.77 &+4.55 &-25.21 &-10.51 &-5.45 \\
            \cmidrule{2-9}
            & \multirow{3}{*}{LoRA$_{(r=32)}$} 
            & Before &52.07\scriptsize{$\pm$0.31} &95.01\scriptsize{$\pm$0.21} &79.23\scriptsize{$\pm$0.22} &52.87\scriptsize{$\pm$1.79} &81.67\scriptsize{$\pm$1.14} &87.80\scriptsize{$\pm$0.72} \\
            & & After &32.86\scriptsize{$\pm$1.06} &55.58\scriptsize{$\pm$0.76} &83.91\scriptsize{$\pm$0.70} &23.84\scriptsize{$\pm$2.13} &66.23\scriptsize{$\pm$1.65} &79.27\scriptsize{$\pm$1.05} \\
            & & $\Delta$ (\%) &-19.21 &-39.43 &+4.68 &-29.03 &-15.44 &-8.53 \\
            \cmidrule{2-9}
            & \multirow{3}{*}{Split-LoRA$_{(4\times8)}$} 
            & Before &50.68\scriptsize{$\pm$1.06} &93.08\scriptsize{$\pm$0.41} &77.12\scriptsize{$\pm$0.76} &48.65\scriptsize{$\pm$1.18} &79.30\scriptsize{$\pm$0.91} &86.05\scriptsize{$\pm$0.44} \\
            & & After &41.83\scriptsize{$\pm$1.92} &59.37\scriptsize{$\pm$0.59} &81.70\scriptsize{$\pm$0.52} &22.98\scriptsize{$\pm$1.35} &67.02\scriptsize{$\pm$0.59} &81.01\scriptsize{$\pm$1.34} \\
            & & $\Delta$ (\%) &-8.85 &-33.71 &+4.58 &-25.67 &-12.28 &-5.04 \\
            \cmidrule{2-9}
            & \multirow{3}{*}{FlyLoRA$_{(k=8)}$} 
            & Before &53.68\scriptsize{$\pm$0.47} &95.55\scriptsize{$\pm$0.18} &80.82\scriptsize{$\pm$0.56} &54.34\scriptsize{$\pm$2.13} &82.85\scriptsize{$\pm$0.52} &89.63\scriptsize{$\pm$0.55} \\
            & & After &60.23\scriptsize{$\pm$0.95} &71.78\scriptsize{$\pm$0.41} &85.62\scriptsize{$\pm$0.43} &33.11\scriptsize{$\pm$1.61} &75.28\scriptsize{$\pm$1.72} &87.15\scriptsize{$\pm$1.26} \\
            & &\cellcolor{gray!20}\textbf{$\Delta$ (\%)} &\cellcolor{gray!20}\textbf{+6.55} &\cellcolor{gray!20}\textbf{-23.77} &\cellcolor{gray!20}\textbf{+4.80} &\cellcolor{gray!20}\textbf{-21.23} &\cellcolor{gray!20}\textbf{-7.57} &\cellcolor{gray!20}\textbf{-2.48} \\
        \midrule
    \end{tabular}}}
    \label{tab:multi_results}
    \vspace{-5pt}
\end{table}

\subsection{Single Task Performance}

The single-task results are presented in Table \ref{tab:single_results}. Despite operating under a lower computational budget, FlyLoRA$_{(k=8)}$ outperforms LoRA variants with the same rank (LoRA$_{(r=8)}$) across all datasets. This improvement can be attributed to its broader parameter space. Notably, FlyLoRA$_{(k=8)}$ also achieves slightly better performance than LoRA variants with the same total rank (LoRA$_{(r=32)}$), suggesting that a significant portion of LoRA's parameters are redundant and may introduce interference. Additionally, FlyLoRA$_{(k=8)}$ demonstrates superior performance over Split-LoRA$_{(4\times8)}$, highlighting the benefits of its finer expert allocation strategy within the MoE framework, which enables \textbf{intra-task decoupling}. The reduction in activated trainable parameters compared to these baselines shows FlyLoRA's efficiency. Extended results with larger models and further baselines are in Appendix \ref{App:additional_results}.

\subsection{Multi-task Performance} \label{sec:multi-task}

For simplicity, we first employ the widely used weight averaging technique for model merging. Specifically, this corresponds to setting $w_i=\frac{1}{t}$ in Eq. \ref{eq:lora_merging}, yielding the merged weights $\bm{W}'=\bm{W}_0 + \frac{1}{t}\sum_{i=1}^t \bm{B}_i\bm{A}_i$. The multi-task results are presented in Table \ref{tab:multi_results}, where LoRA components from different domains are merged. Compared to both LoRA variants ($r=8$ and $r=32$) and Split-LoRA$_{(4\times8)}$, FlyLoRA achieves higher accuracy both before and after merging, with significantly smaller performance degradation. This robustness stems from its \textbf{inter-task decoupling} enabled by approximate orthogonal random projection, as theoretically analyzed in Section \ref{sec:inter_task_orthgonal}. Additional results with advanced fusion techniques are provided in Appendix \ref{App:additional_results}.

\subsection{Ablation Study and Hyperparameter Sensitivity Analysis} \label{sec:ablation_sensitivity}

\begin{table*}[t]
    \vspace{-15pt}
    \centering
    \begin{minipage}{0.42\textwidth}
    \begin{table}[H]
        \centering
        \small
        \caption{Ablation study of FlyLoRA variants analyzing: \textbf{(a)} Load balancing in single-task (ST) setting, \textbf{(b)} Matrix $\bm{A}$ freezing in both ST and multi-task merging (MT) settings, where LB=Load Balancing, Frz=$\bm{A}$ Frozen, Trn=$\bm{A}$ Trainable.}
        \vspace{+2pt}
        \begin{tabular}{@{}lc@{\hskip 1em}lc@{}}
        \toprule
        \multicolumn{2}{c}{\textbf{Load Balancing}} & \multicolumn{2}{c}{\textbf{Frozen $A$}} \\
        \cmidrule(lr){1-2} \cmidrule(lr){3-4}
        Variant & Acc (\%) & Variant & Acc (\%) \\
        \midrule
        w/ LB &\textbf{40.88\scriptsize{$\pm$1.61}} &ST+Frz &\textbf{40.88\scriptsize{$\pm$1.61}} \\
        w/o LB &37.56\scriptsize{$\pm$2.87} &ST+Trn &40.64\scriptsize{$\pm$1.35} \\
        &  &MT+Frz &\textbf{38.86\scriptsize{$\pm$1.46}} \\
        &  &MT+Trn &34.43\scriptsize{$\pm$2.24} \\
        \bottomrule
        \vspace{-10pt}
        \end{tabular}
        \label{tab:ablation}
    \end{table}
    \end{minipage}
    \hfill
    \begin{minipage}{0.54\textwidth}
    \begin{figure}[H]
        \centering
        \begin{subfigure}{0.32\textwidth}
            \includegraphics[width=\textwidth]{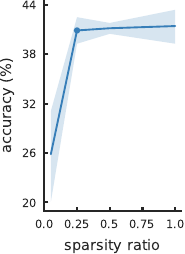}
            \caption{}
        \end{subfigure}
        \begin{subfigure}{0.32\textwidth}
            \includegraphics[width=\textwidth]{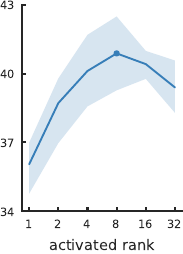}
            \caption{}
        \end{subfigure}
        \begin{subfigure}{0.32\textwidth}
            \includegraphics[width=\textwidth]{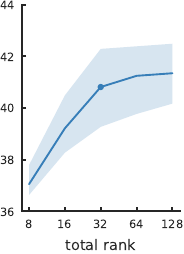}
            \caption{}
        \end{subfigure}
        \caption{Accuracy comparison for: \textbf{(a)} Sparsity ratio in $\bm{A}$, \textbf{(b)} Activated rank (with fixed total rank $r=32$), \textbf{(c)} Total rank (with fixed activated rank $k=8$).}
        \label{fig:sensitivity_analysis}
    \end{figure}
    \end{minipage}
    \vspace{-15pt}
\end{table*}

We conduct an ablation study to analyze key properties of FlyLoRA by evaluating two critical modifications: (1) removing load-balancing strategies and (2) replacing the frozen matrix $\bm{A}$ with an updatable version, as shown in Table \ref{tab:ablation} (MMLU and Llama-3.1-8B). Our results demonstrate that load-balancing significantly improves MoE's training stability and boosts accuracy by $3.32\%$. For matrix $\bm{A}$, we observe minimal performance differences between frozen and updatable versions in single-task settings. However, in multi-task merging scenarios, using an updatable matrix leads to a $4.43\%$ performance degradation, as the updatable $\bm{A}$ does not satisfy the approximately orthogonal property. More ablation studies are provided in Appendix \ref{App:additional_results}.

Further sensitivity analysis on single-task performance (Figure \ref{fig:sensitivity_analysis}) reveals three key insights. First, model accuracy increases monotonically with the sparsity ratio of $\bm{A}$ before saturating, exhibiting only marginal degradation unless the sparsity ratio becomes extremely small. Second, under a fixed total rank budget, performance peaks at intermediate activation ranks: insufficient rank fails to capture task-specific features, while excessive rank induces parameter interference. Third, increasing the total rank while holding the activation rank constant consistently yields performance gains.

\section{Discussion} \label{sec:discussion}

\subsection{Interference in Model Merging}
In scenarios requiring training, gradient orthogonalization techniques are commonly employed to reduce task interference, as seen in multi-task learning \cite{yu2020gradient} and continual learning \cite{chaudhry2020continual, zeng2019continual}. In our training-free model merging setting, all components are derived from the same base model through domain-specific SFT. Task interference can be quantified by measuring the orthogonality of parameter updates (relative to the base model) across different tasks \cite{ilharco2022editing}. For our LoRA components merging, these parameter updates correspond to $\bm{B}_i\bm{A}_i$. We formally prove the near-orthogonality of FlyLoRA in Appendix \ref{App:subspace_orthogonality}, which inherently reduces inter-task correlations.

\subsection{FlyLoRA's Connection to Other Orthogonality-Based Designs in PEFT}
Representative orthogonality-based PEFT methods like OFT \cite{liu2023parameter, qiu2023controlling} and LoReFT \cite{wu2024reft} both operate on single tasks, and their orthogonal matrix $\bm{R}$ multiplies the pre-trained weight matrix $\bm{W}_0$, differing from LoRA variants (including FlyLoRA) that add $\Delta \bm{W}$ to $\bm{W}_0$. The multiplication scheme rotates the entire weight parameter space, and \cite{liu2023parameter, qiu2023controlling} demonstrate this better adjusts semantic information compared to changing magnitude, explaining its success. In contrast, the additive scheme lacks this property since $\bm{W}_0$ cannot be rotated. In single-task settings, removing the MoE part with only random $\bm{A}$ reduces FlyLoRA to LoRA-FA \cite{zhang2023lora} or Asymmetry LoRA \cite{zhu2024asymmetry}. These variants can save resources but cannot improve performance. Thus, although all methods use orthogonality, FlyLoRA succeeds differently. We think the orthogonality design in LoRA excels in \emph{multi-task} scenarios, such as model merging (this work and LoRI \cite{zhang2025lori}) and continual learning (O-LoRA \cite{wang2023orthogonal}), because it decouples parameter interference across multiple downstream tasks when fine-tuning from the base model.

\section{Related Work}
\paragraph{Low-Rank Adaptation}
LoRA \cite{hu2022lora} is a widely used PEFT strategy for fine-tuning LLMs. To enhance its expressive power, several improvements \cite{liu2024dora, zhang2023adalora} have been proposed. Recently, to address parameter interference in settings like multi-task and continual learning, several MoE-based LoRA variants \cite{dou2023loramoe, feng2024mixture, gao2024higher, li2024mixlora, muqeeth2023soft, tian2024hydralora, wu2024mixture, zadouri2023pushing} have proven effective by forcing each expert to specialize in specific areas. Several works \cite{stoica2024model, tang2023parameter, zhang2025lori} also aim to reduce interference during LoRA merging. In this article, we further develop the MoE-based LoRA structure. To improve LoRA's efficiency, LoRA-FA \cite{zhang2023lora} and AsymmetryLoRA \cite{zhu2024asymmetry} show that fixing the down-projection matrix $\bm{A}$ saves memory for input activations without performance degradation. In our work, we reconsider freezing $\bm{A}$ from a new perspective by showing that its orthogonality and distance-preserving properties can be utilized to design an improved intra-/inter-task decoupling mechanism.

\paragraph{Fly Olfactory Circuit}
The fly olfactory circuit \cite{caron2013random, dasgupta2017neural, li2025seemingly, lin2014sparse, stevens2015fly} serves as an exemplary model in bio-inspired AI due to its structural simplicity and functional completeness. Its core mechanism---random projection followed by sparse selection---effectively transforms high-dimensional inputs into separable representations. This biological principle has inspired algorithmic innovations across multiple AI domains, including locality-sensitive hashing \cite{sanjoy2017fly, ryali2020bio, sharma2018improving}, word embedding \cite{liang2021can}, federated learning \cite{ram2022federated}, and continual learning \cite{zou2025structural, zou2025fly}. The circuit's enduring relevance highlights the value of cross-disciplinary inspiration in advancing computational methods.

\section{Conclusion}
In summary, this work provides a comprehensive revisit of the MoE-based structure for LoRA and analyzes its drawbacks regarding parameter interference and efficiency. Inspired by the fly olfactory circuit, we introduce FlyLoRA, a novel MoE-based LoRA variant that employs rank-wise expert activation in matrix $\bm{B}$ and a fixed sparse random projection for matrix $\bm{A}$ as an implicit router. Through the theoretical properties of these components, FlyLoRA achieves both intra-task and inter-task decoupling, significantly improving decorrelation in single-domain instruction tuning and LoRA component fusion in multi-task settings. Additionally, the implicit routing strategy and inherent sparsity ensure computational efficiency.

\section{Acknowledgments}
We thank Cheems Wang for his valuable suggestions on the manuscript. We also thank the anonymous reviewers for their positive feedback and constructive comments. This work was supported by the National Key R\&D Program of China under Grant 2018AAA0102801.

% \newpage
\bibliographystyle{plainnat}
\bibliography{neurips_2025}

\begin{thebibliography}{101}
\providecommand{\natexlab}[1]{#1}
\providecommand{\url}[1]{\texttt{#1}}
\expandafter\ifx\csname urlstyle\endcsname\relax
  \providecommand{\doi}[1]{doi: #1}\else
  \providecommand{\doi}{doi: \begingroup \urlstyle{rm}\Url}\fi

\bibitem[Achiam et~al.(2023)Achiam, Adler, Agarwal, Ahmad, Akkaya, Aleman, Almeida, Altenschmidt, Altman, Anadkat, et~al.]{achiam2023gpt}
Josh Achiam, Steven Adler, Sandhini Agarwal, Lama Ahmad, Ilge Akkaya, Florencia~Leoni Aleman, Diogo Almeida, Janko Altenschmidt, Sam Altman, Shyamal Anadkat, et~al.
\newblock Gpt-4 technical report.
\newblock \emph{arXiv preprint arXiv:2303.08774}, 2023.

\bibitem[Aghajanyan et~al.(2020)Aghajanyan, Zettlemoyer, and Gupta]{aghajanyan2020intrinsic}
Armen Aghajanyan, Luke Zettlemoyer, and Sonal Gupta.
\newblock Intrinsic dimensionality explains the effectiveness of language model fine-tuning.
\newblock \emph{arXiv preprint arXiv:2012.13255}, 2020.

\bibitem[Ailon and Chazelle(2009)]{ailon2009fast}
Nir Ailon and Bernard Chazelle.
\newblock The fast johnson--lindenstrauss transform and approximate nearest neighbors.
\newblock \emph{SIAM Journal on computing}, 39\penalty0 (1):\penalty0 302--322, 2009.

\bibitem[Brown et~al.(2020)Brown, Mann, Ryder, Subbiah, Kaplan, Dhariwal, Neelakantan, Shyam, Sastry, Askell, et~al.]{brown2020language}
Tom Brown, Benjamin Mann, Nick Ryder, Melanie Subbiah, Jared~D Kaplan, Prafulla Dhariwal, Arvind Neelakantan, Pranav Shyam, Girish Sastry, Amanda Askell, et~al.
\newblock Language models are few-shot learners.
\newblock \emph{Advances in neural information processing systems}, 33:\penalty0 1877--1901, 2020.

\bibitem[Bruhin and Davies(2022)]{bruhin2022bioinspired}
Nina~Dekoninck Bruhin and Bryn Davies.
\newblock Bioinspired random projections for robust, sparse classification.
\newblock \emph{SIAM Journal on Imaging Sciences}, 15\penalty0 (4):\penalty0 1833--1850, 2022.

\bibitem[Caron et~al.(2013)Caron, Ruta, Abbott, and Axel]{caron2013random}
Sophie~JC Caron, Vanessa Ruta, Larry~F Abbott, and Richard Axel.
\newblock Random convergence of olfactory inputs in the drosophila mushroom body.
\newblock \emph{Nature}, 497\penalty0 (7447):\penalty0 113--117, 2013.

\bibitem[Chaudhary(2023)]{codealpaca}
Sahil Chaudhary.
\newblock Code alpaca: An instruction-following llama model for code generation.
\newblock \url{https://github.com/sahil280114/codealpaca}, 2023.

\bibitem[Chaudhry et~al.(2020)Chaudhry, Khan, Dokania, and Torr]{chaudhry2020continual}
Arslan Chaudhry, Naeemullah Khan, Puneet Dokania, and Philip Torr.
\newblock Continual learning in low-rank orthogonal subspaces.
\newblock \emph{Advances in Neural Information Processing Systems}, 33:\penalty0 9900--9911, 2020.

\bibitem[Chen et~al.(2021)Chen, Tworek, Jun, Yuan, de~Oliveira~Pinto, Kaplan, Edwards, Burda, Joseph, Brockman, Ray, Puri, Krueger, Petrov, Khlaaf, Sastry, Mishkin, Chan, Gray, Ryder, Pavlov, Power, Kaiser, Bavarian, Winter, Tillet, Such, Cummings, Plappert, Chantzis, Barnes, Herbert-Voss, Guss, Nichol, Paino, Tezak, Tang, Babuschkin, Balaji, Jain, Saunders, Hesse, Carr, Leike, Achiam, Misra, Morikawa, Radford, Knight, Brundage, Murati, Mayer, Welinder, McGrew, Amodei, McCandlish, Sutskever, and Zaremba]{chen2021evaluating}
Mark Chen, Jerry Tworek, Heewoo Jun, Qiming Yuan, Henrique~Ponde de~Oliveira~Pinto, Jared Kaplan, Harri Edwards, Yuri Burda, Nicholas Joseph, Greg Brockman, Alex Ray, Raul Puri, Gretchen Krueger, Michael Petrov, Heidy Khlaaf, Girish Sastry, Pamela Mishkin, Brooke Chan, Scott Gray, Nick Ryder, Mikhail Pavlov, Alethea Power, Lukasz Kaiser, Mohammad Bavarian, Clemens Winter, Philippe Tillet, Felipe~Petroski Such, Dave Cummings, Matthias Plappert, Fotios Chantzis, Elizabeth Barnes, Ariel Herbert-Voss, William~Hebgen Guss, Alex Nichol, Alex Paino, Nikolas Tezak, Jie Tang, Igor Babuschkin, Suchir Balaji, Shantanu Jain, William Saunders, Christopher Hesse, Andrew~N. Carr, Jan Leike, Josh Achiam, Vedant Misra, Evan Morikawa, Alec Radford, Matthew Knight, Miles Brundage, Mira Murati, Katie Mayer, Peter Welinder, Bob McGrew, Dario Amodei, Sam McCandlish, Ilya Sutskever, and Wojciech Zaremba.
\newblock Evaluating large language models trained on code, 2021.

\bibitem[Chowdhury et~al.(2023)Chowdhury, Zhang, Wang, Liu, and Chen]{chowdhury2023patch}
Mohammed Nowaz~Rabbani Chowdhury, Shuai Zhang, Meng Wang, Sijia Liu, and Pin-Yu Chen.
\newblock Patch-level routing in mixture-of-experts is provably sample-efficient for convolutional neural networks.
\newblock In \emph{International Conference on Machine Learning}, pages 6074--6114. PMLR, 2023.

\bibitem[Chronopoulou et~al.(2023)Chronopoulou, Peters, Fraser, and Dodge]{chronopoulou2023adaptersoup}
Alexandra Chronopoulou, Matthew~E Peters, Alexander Fraser, and Jesse Dodge.
\newblock Adaptersoup: Weight averaging to improve generalization of pretrained language models.
\newblock \emph{arXiv preprint arXiv:2302.07027}, 2023.

\bibitem[Cobbe et~al.(2021)Cobbe, Kosaraju, Bavarian, Chen, Jun, Kaiser, Plappert, Tworek, Hilton, Nakano, Hesse, and Schulman]{cobbe2021gsm8k}
Karl Cobbe, Vineet Kosaraju, Mohammad Bavarian, Mark Chen, Heewoo Jun, Lukasz Kaiser, Matthias Plappert, Jerry Tworek, Jacob Hilton, Reiichiro Nakano, Christopher Hesse, and John Schulman.
\newblock Training verifiers to solve math word problems.
\newblock \emph{arXiv preprint arXiv:2110.14168}, 2021.

\bibitem[Dasgupta et~al.(2017{\natexlab{a}})Dasgupta, Stevens, and Navlakha]{dasgupta2017neural}
Sanjoy Dasgupta, Charles~F Stevens, and Saket Navlakha.
\newblock A neural algorithm for a fundamental computing problem.
\newblock \emph{Science}, 358\penalty0 (6364):\penalty0 793--796, 2017{\natexlab{a}}.

\bibitem[Dasgupta et~al.(2017{\natexlab{b}})Dasgupta, Stevens, and Navlakha]{sanjoy2017fly}
Sanjoy Dasgupta, Charles~F. Stevens, and Saket Navlakha.
\newblock A neural algorithm for a fundamental computing problem.
\newblock \emph{Science}, 358\penalty0 (6364):\penalty0 793--796, 2017{\natexlab{b}}.
\newblock \doi{10.1126/science.aam9868}.

\bibitem[Devlin et~al.(2019)Devlin, Chang, Lee, and Toutanova]{devlin2019bert}
Jacob Devlin, Ming-Wei Chang, Kenton Lee, and Kristina Toutanova.
\newblock Bert: Pre-training of deep bidirectional transformers for language understanding.
\newblock In \emph{Proceedings of the 2019 conference of the North American chapter of the association for computational linguistics: human language technologies, volume 1 (long and short papers)}, pages 4171--4186, 2019.

\bibitem[Ding et~al.(2023)Ding, Lv, Wang, Chen, Zhou, Liu, and Sun]{ding2023sparse}
Ning Ding, Xingtai Lv, Qiaosen Wang, Yulin Chen, Bowen Zhou, Zhiyuan Liu, and Maosong Sun.
\newblock Sparse low-rank adaptation of pre-trained language models.
\newblock \emph{arXiv preprint arXiv:2311.11696}, 2023.

\bibitem[Dou et~al.(2023)Dou, Zhou, Liu, Gao, Zhao, Shen, Zhou, Xi, Wang, Fan, et~al.]{dou2023loramoe}
Shihan Dou, Enyu Zhou, Yan Liu, Songyang Gao, Jun Zhao, Wei Shen, Yuhao Zhou, Zhiheng Xi, Xiao Wang, Xiaoran Fan, et~al.
\newblock Loramoe: Alleviate world knowledge forgetting in large language models via moe-style plugin.
\newblock \emph{arXiv preprint arXiv:2312.09979}, 2023.

\bibitem[Eckart and Young(1936)]{eckart1936approximation}
Carl Eckart and Gale Young.
\newblock The approximation of one matrix by another of lower rank.
\newblock \emph{Psychometrika}, 1\penalty0 (3):\penalty0 211--218, 1936.

\bibitem[Fedus et~al.(2022)Fedus, Zoph, and Shazeer]{fedus2022switch}
William Fedus, Barret Zoph, and Noam Shazeer.
\newblock Switch transformers: Scaling to trillion parameter models with simple and efficient sparsity.
\newblock \emph{Journal of Machine Learning Research}, 23\penalty0 (120):\penalty0 1--39, 2022.

\bibitem[Feng et~al.(2024)Feng, Hao, Zhang, Han, and Wang]{feng2024mixture}
Wenfeng Feng, Chuzhan Hao, Yuewei Zhang, Yu~Han, and Hao Wang.
\newblock Mixture-of-loras: An efficient multitask tuning for large language models.
\newblock \emph{arXiv preprint arXiv:2403.03432}, 2024.

\bibitem[Gao et~al.(2024)Gao, Chen, Rao, Sun, Liu, Peng, Zhang, Guo, Yang, and Subrahmanian]{gao2024higher}
Chongyang Gao, Kezhen Chen, Jinmeng Rao, Baochen Sun, Ruibo Liu, Daiyi Peng, Yawen Zhang, Xiaoyuan Guo, Jie Yang, and VS~Subrahmanian.
\newblock Higher layers need more lora experts.
\newblock \emph{arXiv preprint arXiv:2402.08562}, 2024.

\bibitem[Gekhman et~al.(2024)Gekhman, Yona, Aharoni, Eyal, Feder, Reichart, and Herzig]{gekhman2024does}
Zorik Gekhman, Gal Yona, Roee Aharoni, Matan Eyal, Amir Feder, Roi Reichart, and Jonathan Herzig.
\newblock Does fine-tuning llms on new knowledge encourage hallucinations?
\newblock \emph{arXiv preprint arXiv:2405.05904}, 2024.

\bibitem[Grattafiori et~al.(2024)Grattafiori, Dubey, Jauhri, Pandey, Kadian, Al-Dahle, Letman, Mathur, Schelten, Vaughan, et~al.]{grattafiori2024llama}
Aaron Grattafiori, Abhimanyu Dubey, Abhinav Jauhri, Abhinav Pandey, Abhishek Kadian, Ahmad Al-Dahle, Aiesha Letman, Akhil Mathur, Alan Schelten, Alex Vaughan, et~al.
\newblock The llama 3 herd of models.
\newblock \emph{arXiv preprint arXiv:2407.21783}, 2024.

\bibitem[Guo et~al.(2025)Guo, Yang, Zhang, Song, Zhang, Xu, Zhu, Ma, Wang, Bi, et~al.]{guo2025deepseek}
Daya Guo, Dejian Yang, Haowei Zhang, Junxiao Song, Ruoyu Zhang, Runxin Xu, Qihao Zhu, Shirong Ma, Peiyi Wang, Xiao Bi, et~al.
\newblock Deepseek-r1: Incentivizing reasoning capability in llms via reinforcement learning.
\newblock \emph{arXiv preprint arXiv:2501.12948}, 2025.

\bibitem[Hendrycks et~al.(2021)Hendrycks, Burns, Basart, Zou, Mazeika, Song, and Steinhardt]{hendryckstest2021}
Dan Hendrycks, Collin Burns, Steven Basart, Andy Zou, Mantas Mazeika, Dawn Song, and Jacob Steinhardt.
\newblock Measuring massive multitask language understanding.
\newblock \emph{Proceedings of the International Conference on Learning Representations (ICLR)}, 2021.

\bibitem[Houlsby et~al.(2019)Houlsby, Giurgiu, Jastrzebski, Morrone, De~Laroussilhe, Gesmundo, Attariyan, and Gelly]{houlsby2019parameter}
Neil Houlsby, Andrei Giurgiu, Stanislaw Jastrzebski, Bruna Morrone, Quentin De~Laroussilhe, Andrea Gesmundo, Mona Attariyan, and Sylvain Gelly.
\newblock Parameter-efficient transfer learning for nlp.
\newblock In \emph{International conference on machine learning}, pages 2790--2799. PMLR, 2019.

\bibitem[Hu et~al.(2022)Hu, Shen, Wallis, Allen-Zhu, Li, Wang, Wang, Chen, et~al.]{hu2022lora}
Edward~J Hu, Yelong Shen, Phillip Wallis, Zeyuan Allen-Zhu, Yuanzhi Li, Shean Wang, Lu~Wang, Weizhu Chen, et~al.
\newblock Lora: Low-rank adaptation of large language models.
\newblock \emph{ICLR}, 1\penalty0 (2):\penalty0 3, 2022.

\bibitem[Huang et~al.(2023)Huang, Liu, Lin, Pang, Du, and Lin]{huang2023lorahub}
Chengsong Huang, Qian Liu, Bill~Yuchen Lin, Tianyu Pang, Chao Du, and Min Lin.
\newblock Lorahub: Efficient cross-task generalization via dynamic lora composition.
\newblock \emph{arXiv preprint arXiv:2307.13269}, 2023.

\bibitem[Ilharco et~al.(2022)Ilharco, Ribeiro, Wortsman, Gururangan, Schmidt, Hajishirzi, and Farhadi]{ilharco2022editing}
Gabriel Ilharco, Marco~Tulio Ribeiro, Mitchell Wortsman, Suchin Gururangan, Ludwig Schmidt, Hannaneh Hajishirzi, and Ali Farhadi.
\newblock Editing models with task arithmetic.
\newblock \emph{arXiv preprint arXiv:2212.04089}, 2022.

\bibitem[Jacobs et~al.(1991)Jacobs, Jordan, Nowlan, and Hinton]{jacobs1991adaptive}
Robert~A Jacobs, Michael~I Jordan, Steven~J Nowlan, and Geoffrey~E Hinton.
\newblock Adaptive mixtures of local experts.
\newblock \emph{Neural computation}, 3\penalty0 (1):\penalty0 79--87, 1991.

\bibitem[Jiang et~al.(2024)Jiang, Huang, Luo, Zhang, Huang, Wei, Deng, Sun, Zhang, Wang, et~al.]{jiang2024mora}
Ting Jiang, Shaohan Huang, Shengyue Luo, Zihan Zhang, Haizhen Huang, Furu Wei, Weiwei Deng, Feng Sun, Qi~Zhang, Deqing Wang, et~al.
\newblock Mora: High-rank updating for parameter-efficient fine-tuning.
\newblock \emph{arXiv preprint arXiv:2405.12130}, 2024.

\bibitem[Johnson et~al.(1984)Johnson, Lindenstrauss, et~al.]{johnson1984extensions}
William~B Johnson, Joram Lindenstrauss, et~al.
\newblock Extensions of lipschitz mappings into a hilbert space.
\newblock \emph{Contemporary mathematics}, 26\penalty0 (189-206):\penalty0 1, 1984.

\bibitem[Kornblith et~al.(2019)Kornblith, Norouzi, Lee, and Hinton]{kornblith2019similarity}
Simon Kornblith, Mohammad Norouzi, Honglak Lee, and Geoffrey Hinton.
\newblock Similarity of neural network representations revisited.
\newblock In \emph{International conference on machine learning}, pages 3519--3529. PMlR, 2019.

\bibitem[Lepikhin et~al.(2020)Lepikhin, Lee, Xu, Chen, Firat, Huang, Krikun, Shazeer, and Chen]{lepikhin2020gshard}
Dmitry Lepikhin, HyoukJoong Lee, Yuanzhong Xu, Dehao Chen, Orhan Firat, Yanping Huang, Maxim Krikun, Noam Shazeer, and Zhifeng Chen.
\newblock Gshard: Scaling giant models with conditional computation and automatic sharding.
\newblock \emph{arXiv preprint arXiv:2006.16668}, 2020.

\bibitem[Lester et~al.(2021)Lester, Al-Rfou, and Constant]{lester2021power}
Brian Lester, Rami Al-Rfou, and Noah Constant.
\newblock The power of scale for parameter-efficient prompt tuning.
\newblock \emph{arXiv preprint arXiv:2104.08691}, 2021.

\bibitem[Li et~al.(2018)Li, Farkhoor, Liu, and Yosinski]{li2018measuring}
Chunyuan Li, Heerad Farkhoor, Rosanne Liu, and Jason Yosinski.
\newblock Measuring the intrinsic dimension of objective landscapes.
\newblock \emph{arXiv preprint arXiv:1804.08838}, 2018.

\bibitem[Li et~al.(2024)Li, Ma, Wang, Ye, Cheng, Tang, Zhang, Duan, Zuo, Yang, et~al.]{li2024mixlora}
Dengchun Li, Yingzi Ma, Naizheng Wang, Zhengmao Ye, Zhiyuan Cheng, Yinghao Tang, Yan Zhang, Lei Duan, Jie Zuo, Cal Yang, et~al.
\newblock Mixlora: Enhancing large language models fine-tuning with lora-based mixture of experts.
\newblock \emph{arXiv preprint arXiv:2404.15159}, 2024.

\bibitem[Li et~al.(2025{\natexlab{a}})Li, Yu, Yu, and Zang]{li2025seemingly}
Haiyang Li, Liao Yu, Qiang Yu, and Yunliang Zang.
\newblock Seemingly redundant modules enhance robust odor learning in fruit flies.
\newblock In \emph{The Thirty-ninth Annual Conference on Neural Information Processing Systems}, 2025{\natexlab{a}}.

\bibitem[Li et~al.(2025{\natexlab{b}})Li, Zhang, Zhang, Wang, Liu, and Chen]{li2025task}
Hongkang Li, Yihua Zhang, Shuai Zhang, Meng Wang, Sijia Liu, and Pin-Yu Chen.
\newblock When is task vector provably effective for model editing? a generalization analysis of nonlinear transformers.
\newblock \emph{arXiv preprint arXiv:2504.10957}, 2025{\natexlab{b}}.

\bibitem[Li and Liang(2021)]{li2021prefix}
Xiang~Lisa Li and Percy Liang.
\newblock Prefix-tuning: Optimizing continuous prompts for generation.
\newblock \emph{arXiv preprint arXiv:2101.00190}, 2021.

\bibitem[Liang et~al.(2021)Liang, Ryali, Hoover, Grinberg, Navlakha, Zaki, and Krotov]{liang2021can}
Yuchen Liang, Chaitanya~K Ryali, Benjamin Hoover, Leopold Grinberg, Saket Navlakha, Mohammed~J Zaki, and Dmitry Krotov.
\newblock Can a fruit fly learn word embeddings?
\newblock \emph{arXiv preprint arXiv:2101.06887}, 2021.

\bibitem[Lin et~al.(2014)Lin, Bygrave, De~Calignon, Lee, and Miesenb{\"o}ck]{lin2014sparse}
Andrew~C Lin, Alexei~M Bygrave, Alix De~Calignon, Tzumin Lee, and Gero Miesenb{\"o}ck.
\newblock Sparse, decorrelated odor coding in the mushroom body enhances learned odor discrimination.
\newblock \emph{Nature neuroscience}, 17\penalty0 (4):\penalty0 559--568, 2014.

\bibitem[Liu et~al.(2024{\natexlab{a}})Liu, Feng, Xue, Wang, Wu, Lu, Zhao, Deng, Zhang, Ruan, et~al.]{liu2024deepseek}
Aixin Liu, Bei Feng, Bing Xue, Bingxuan Wang, Bochao Wu, Chengda Lu, Chenggang Zhao, Chengqi Deng, Chenyu Zhang, Chong Ruan, et~al.
\newblock Deepseek-v3 technical report.
\newblock \emph{arXiv preprint arXiv:2412.19437}, 2024{\natexlab{a}}.

\bibitem[Liu et~al.(2022)Liu, Tam, Muqeeth, Mohta, Huang, Bansal, and Raffel]{liu2022few}
Haokun Liu, Derek Tam, Mohammed Muqeeth, Jay Mohta, Tenghao Huang, Mohit Bansal, and Colin~A Raffel.
\newblock Few-shot parameter-efficient fine-tuning is better and cheaper than in-context learning.
\newblock \emph{Advances in Neural Information Processing Systems}, 35:\penalty0 1950--1965, 2022.

\bibitem[Liu et~al.(2024{\natexlab{b}})Liu, Wang, Yin, Molchanov, Wang, Cheng, and Chen]{liu2024dora}
Shih-Yang Liu, Chien-Yi Wang, Hongxu Yin, Pavlo Molchanov, Yu-Chiang~Frank Wang, Kwang-Ting Cheng, and Min-Hung Chen.
\newblock Dora: Weight-decomposed low-rank adaptation.
\newblock In \emph{Forty-first International Conference on Machine Learning}, 2024{\natexlab{b}}.

\bibitem[Liu et~al.(2023)Liu, Qiu, Feng, Xiu, Xue, Yu, Feng, Liu, Heo, Peng, et~al.]{liu2023parameter}
Weiyang Liu, Zeju Qiu, Yao Feng, Yuliang Xiu, Yuxuan Xue, Longhui Yu, Haiwen Feng, Zhen Liu, Juyeon Heo, Songyou Peng, et~al.
\newblock Parameter-efficient orthogonal finetuning via butterfly factorization.
\newblock \emph{arXiv preprint arXiv:2311.06243}, 2023.

\bibitem[Liu et~al.(2024{\natexlab{c}})Liu, Zheng, Du, Ding, Qian, Yang, and Tang]{liu2024gpt}
Xiao Liu, Yanan Zheng, Zhengxiao Du, Ming Ding, Yujie Qian, Zhilin Yang, and Jie Tang.
\newblock Gpt understands, too.
\newblock \emph{AI Open}, 5:\penalty0 208--215, 2024{\natexlab{c}}.

\bibitem[Lu et~al.(2022)Lu, Mishra, Xia, Qiu, Chang, Zhu, Tafjord, Clark, and Kalyan]{lu2022learn}
Pan Lu, Swaroop Mishra, Tony Xia, Liang Qiu, Kai-Wei Chang, Song-Chun Zhu, Oyvind Tafjord, Peter Clark, and Ashwin Kalyan.
\newblock Learn to explain: Multimodal reasoning via thought chains for science question answering.
\newblock In \emph{The 36th Conference on Neural Information Processing Systems (NeurIPS)}, 2022.

\bibitem[Lv et~al.(2025)Lv, Xie, Qian, Wu, Sun, Kang, Wang, and Yan]{lv2025autonomy}
Ang Lv, Ruobing Xie, Yining Qian, Songhao Wu, Xingwu Sun, Zhanhui Kang, Di~Wang, and Rui Yan.
\newblock Autonomy-of-experts models.
\newblock \emph{arXiv preprint arXiv:2501.13074}, 2025.

\bibitem[Mao et~al.(2023{\natexlab{a}})Mao, Zhang, Chen, Xu, and Ji]{mao2023supporteda}
Yixiu Mao, Hongchang Zhang, Chen Chen, Yi~Xu, and Xiangyang Ji.
\newblock Supported trust region optimization for offline reinforcement learning.
\newblock In \emph{International Conference on Machine Learning}, pages 23829--23851. PMLR, 2023{\natexlab{a}}.

\bibitem[Mao et~al.(2023{\natexlab{b}})Mao, Zhang, Chen, Xu, and Ji]{mao2023supportedb}
Yixiu Mao, Hongchang Zhang, Chen Chen, Yi~Xu, and Xiangyang Ji.
\newblock Supported value regularization for offline reinforcement learning.
\newblock \emph{Advances in Neural Information Processing Systems}, 36:\penalty0 40587--40609, 2023{\natexlab{b}}.

\bibitem[Mao et~al.(2024{\natexlab{a}})Mao, Wang, Chen, Qu, and Ji]{mao2024offline}
Yixiu Mao, Qi~Wang, Chen Chen, Yun Qu, and Xiangyang Ji.
\newblock Offline reinforcement learning with ood state correction and ood action suppression.
\newblock \emph{Advances in Neural Information Processing Systems}, 37:\penalty0 93568--93601, 2024{\natexlab{a}}.

\bibitem[Mao et~al.(2024{\natexlab{b}})Mao, Wang, Qu, Jiang, and Ji]{mao2024doubly}
Yixiu Mao, Qi~Wang, Yun Qu, Yuhang Jiang, and Xiangyang Ji.
\newblock Doubly mild generalization for offline reinforcement learning.
\newblock \emph{Advances in Neural Information Processing Systems}, 37:\penalty0 51436--51473, 2024{\natexlab{b}}.

\bibitem[Matena and Raffel(2022)]{matena2022merging}
Michael~S Matena and Colin~A Raffel.
\newblock Merging models with fisher-weighted averaging.
\newblock \emph{Advances in Neural Information Processing Systems}, 35:\penalty0 17703--17716, 2022.

\bibitem[Muqeeth et~al.(2023)Muqeeth, Liu, and Raffel]{muqeeth2023soft}
Mohammed Muqeeth, Haokun Liu, and Colin Raffel.
\newblock Soft merging of experts with adaptive routing.
\newblock \emph{arXiv preprint arXiv:2306.03745}, 2023.

\bibitem[Ortiz-Jimenez et~al.(2023)Ortiz-Jimenez, Favero, and Frossard]{ortiz2023task}
Guillermo Ortiz-Jimenez, Alessandro Favero, and Pascal Frossard.
\newblock Task arithmetic in the tangent space: Improved editing of pre-trained models.
\newblock \emph{Advances in Neural Information Processing Systems}, 36:\penalty0 66727--66754, 2023.

\bibitem[Ouyang et~al.(2022)Ouyang, Wu, Jiang, Almeida, Wainwright, Mishkin, Zhang, Agarwal, Slama, Ray, et~al.]{ouyang2022training}
Long Ouyang, Jeffrey Wu, Xu~Jiang, Diogo Almeida, Carroll Wainwright, Pamela Mishkin, Chong Zhang, Sandhini Agarwal, Katarina Slama, Alex Ray, et~al.
\newblock Training language models to follow instructions with human feedback.
\newblock \emph{Advances in neural information processing systems}, 35:\penalty0 27730--27744, 2022.

\bibitem[Pennington et~al.(2017)Pennington, Schoenholz, and Ganguli]{pennington2017resurrecting}
Jeffrey Pennington, Samuel Schoenholz, and Surya Ganguli.
\newblock Resurrecting the sigmoid in deep learning through dynamical isometry: theory and practice.
\newblock \emph{Advances in neural information processing systems}, 30, 2017.

\bibitem[Qiu et~al.(2023)Qiu, Liu, Feng, Xue, Feng, Liu, Zhang, Weller, and Sch{\"o}lkopf]{qiu2023controlling}
Zeju Qiu, Weiyang Liu, Haiwen Feng, Yuxuan Xue, Yao Feng, Zhen Liu, Dan Zhang, Adrian Weller, and Bernhard Sch{\"o}lkopf.
\newblock Controlling text-to-image diffusion by orthogonal finetuning.
\newblock \emph{Advances in Neural Information Processing Systems}, 36:\penalty0 79320--79362, 2023.

\bibitem[Qu et~al.(2023)Qu, Wang, Shao, Jiang, Chen, Ye, Linc, Feng, Lai, Qin, et~al.]{qu2023hokoff}
Yun Qu, Boyuan Wang, Jianzhun Shao, Yuhang Jiang, Chen Chen, Zhenbin Ye, Liu Linc, Yang Feng, Lin Lai, Hongyang Qin, et~al.
\newblock Hokoff: Real game dataset from honor of kings and its offline reinforcement learning benchmarks.
\newblock \emph{Advances in Neural Information Processing Systems}, 36:\penalty0 22166--22190, 2023.

\bibitem[Qu et~al.(2025{\natexlab{a}})Qu, Wang, Mao, Hu, Ommer, and Ji]{qu2025can}
Yun Qu, Qi~Wang, Yixiu Mao, Vincent~Tao Hu, Bj{\"o}rn Ommer, and Xiangyang Ji.
\newblock Can prompt difficulty be online predicted for accelerating rl finetuning of reasoning models?
\newblock \emph{arXiv preprint arXiv:2507.04632}, 2025{\natexlab{a}}.

\bibitem[Qu et~al.(2025{\natexlab{b}})Qu, Wang, Mao, Lv, and Ji]{qu2025fast}
Yun Qu, Qi~Cheems Wang, Yixiu Mao, Yiqin Lv, and Xiangyang Ji.
\newblock Fast and robust: Task sampling with posterior and diversity synergies for adaptive decision-makers in randomized environments.
\newblock \emph{arXiv preprint arXiv:2504.19139}, 2025{\natexlab{b}}.

\bibitem[Radford et~al.(2018)Radford, Narasimhan, Salimans, Sutskever, et~al.]{radford2018improving}
Alec Radford, Karthik Narasimhan, Tim Salimans, Ilya Sutskever, et~al.
\newblock Improving language understanding by generative pre-training.
\newblock 2018.

\bibitem[Radford et~al.(2019)Radford, Wu, Child, Luan, Amodei, Sutskever, et~al.]{radford2019language}
Alec Radford, Jeffrey Wu, Rewon Child, David Luan, Dario Amodei, Ilya Sutskever, et~al.
\newblock Language models are unsupervised multitask learners.
\newblock \emph{OpenAI blog}, 1\penalty0 (8):\penalty0 9, 2019.

\bibitem[Ram and Sinha(2022)]{ram2022federated}
Parikshit Ram and Kaushik Sinha.
\newblock Federated nearest neighbor classification with a colony of fruit-flies.
\newblock In \emph{Proceedings of the AAAI Conference on Artificial Intelligence}, volume~36, pages 8036--8044, 2022.

\bibitem[Roller et~al.(2021)Roller, Sukhbaatar, Weston, et~al.]{roller2021hash}
Stephen Roller, Sainbayar Sukhbaatar, Jason Weston, et~al.
\newblock Hash layers for large sparse models.
\newblock \emph{advances in neural information processing systems}, 34:\penalty0 17555--17566, 2021.

\bibitem[Ryali et~al.(2020)Ryali, Hopfield, Grinberg, and Krotov]{ryali2020bio}
Chaitanya Ryali, John Hopfield, Leopold Grinberg, and Dmitry Krotov.
\newblock Bio-inspired hashing for unsupervised similarity search.
\newblock In \emph{International conference on machine learning}, pages 8295--8306. PMLR, 2020.

\bibitem[Shao et~al.(2023)Shao, Qu, Chen, Zhang, and Ji]{shao2023counterfactual}
Jianzhun Shao, Yun Qu, Chen Chen, Hongchang Zhang, and Xiangyang Ji.
\newblock Counterfactual conservative q learning for offline multi-agent reinforcement learning.
\newblock \emph{Advances in Neural Information Processing Systems}, 36:\penalty0 77290--77312, 2023.

\bibitem[Sharma and Navlakha(2018)]{sharma2018improving}
Jaiyam Sharma and Saket Navlakha.
\newblock Improving similarity search with high-dimensional locality-sensitive hashing.
\newblock \emph{arXiv preprint arXiv:1812.01844}, 2018.

\bibitem[Shazeer et~al.(2017)Shazeer, Mirhoseini, Maziarz, Davis, Le, Hinton, and Dean]{shazeer2017outrageously}
Noam Shazeer, Azalia Mirhoseini, Krzysztof Maziarz, Andy Davis, Quoc Le, Geoffrey Hinton, and Jeff Dean.
\newblock Outrageously large neural networks: The sparsely-gated mixture-of-experts layer.
\newblock \emph{arXiv preprint arXiv:1701.06538}, 2017.

\bibitem[Shen et~al.(2024)Shen, Wang, Xiao, Van~Noord, and Worring]{shen2024go4align}
Jiayi Shen, Qi~Wang, Zehao Xiao, Nanne Van~Noord, and Marcel Worring.
\newblock Go4align: Group optimization for multi-task alignment.
\newblock \emph{Advances in Neural Information Processing Systems}, 37:\penalty0 111382--111405, 2024.

\bibitem[Stevens(2015)]{stevens2015fly}
Charles~F Stevens.
\newblock What the fly’s nose tells the fly’s brain.
\newblock \emph{Proceedings of the National Academy of Sciences}, 112\penalty0 (30):\penalty0 9460--9465, 2015.

\bibitem[Stoica et~al.(2024)Stoica, Ramesh, Ecsedi, Choshen, and Hoffman]{stoica2024model}
George Stoica, Pratik Ramesh, Boglarka Ecsedi, Leshem Choshen, and Judy Hoffman.
\newblock Model merging with svd to tie the knots.
\newblock \emph{arXiv preprint arXiv:2410.19735}, 2024.

\bibitem[Sun et~al.(2024)Sun, Zhu, Zhang, Yan, Wang, and Ji]{sun2024unleashing}
Peng Sun, Yao Zhu, Yunjian Zhang, Xiu Yan, Zizhe Wang, and Xiangyang Ji.
\newblock Unleashing the potential of large language models through spectral modulation.
\newblock In \emph{Findings of the Association for Computational Linguistics: EMNLP 2024}, pages 3892--3911, 2024.

\bibitem[Tang et~al.(2023)Tang, Shen, Luo, Zhan, Hu, Du, Chen, and Tao]{tang2023parameter}
Anke Tang, Li~Shen, Yong Luo, Yibing Zhan, Han Hu, Bo~Du, Yixin Chen, and Dacheng Tao.
\newblock Parameter efficient multi-task model fusion with partial linearization.
\newblock \emph{arXiv preprint arXiv:2310.04742}, 2023.

\bibitem[Tian et~al.(2024)Tian, Shi, Guo, Li, and Xu]{tian2024hydralora}
Chunlin Tian, Zhan Shi, Zhijiang Guo, Li~Li, and Cheng-Zhong Xu.
\newblock Hydralora: An asymmetric lora architecture for efficient fine-tuning.
\newblock \emph{Advances in Neural Information Processing Systems}, 37:\penalty0 9565--9584, 2024.

\bibitem[Wang et~al.(2024{\natexlab{a}})Wang, Wu, Song, Mittal, and Jia]{wang2024greats}
Jiachen~Tianhao Wang, Tong Wu, Dawn Song, Prateek Mittal, and Ruoxi Jia.
\newblock Greats: Online selection of high-quality data for llm training in every iteration.
\newblock \emph{Advances in Neural Information Processing Systems}, 37:\penalty0 131197--131223, 2024{\natexlab{a}}.

\bibitem[Wang and Van~Hoof(2022)]{wang2022learning}
Qi~Wang and Herke Van~Hoof.
\newblock Learning expressive meta-representations with mixture of expert neural processes.
\newblock \emph{Advances in neural information processing systems}, 35:\penalty0 26242--26255, 2022.

\bibitem[Wang et~al.(2025)Wang, Xiao, Mao, Qu, Shen, Lv, and Ji]{wang2025model}
Qi~Cheems Wang, Zehao Xiao, Yixiu Mao, Yun Qu, Jiayi Shen, Yiqin Lv, and Xiangyang Ji.
\newblock Model predictive task sampling for efficient and robust adaptation.
\newblock \emph{arXiv preprint arXiv:2501.11039}, 2025.

\bibitem[Wang et~al.(2023)Wang, Chen, Ge, Xia, Bao, Zheng, Zhang, Gui, and Huang]{wang2023orthogonal}
Xiao Wang, Tianze Chen, Qiming Ge, Han Xia, Rong Bao, Rui Zheng, Qi~Zhang, Tao Gui, and Xuanjing Huang.
\newblock Orthogonal subspace learning for language model continual learning.
\newblock \emph{arXiv preprint arXiv:2310.14152}, 2023.

\bibitem[Wang et~al.(2024{\natexlab{b}})Wang, Che, Wang, Li, Shi, and Wang]{wang2024separable}
Ziqi Wang, Chang Che, Qi~Wang, Yangyang Li, Zenglin Shi, and Meng Wang.
\newblock Separable mixture of low-rank adaptation for continual visual instruction tuning.
\newblock \emph{arXiv preprint arXiv:2411.13949}, 2024{\natexlab{b}}.

\bibitem[Wu et~al.(2024{\natexlab{a}})Wu, Huang, and Wei]{wu2024mixture}
Xun Wu, Shaohan Huang, and Furu Wei.
\newblock Mixture of lora experts.
\newblock \emph{arXiv preprint arXiv:2404.13628}, 2024{\natexlab{a}}.

\bibitem[Wu et~al.(2024{\natexlab{b}})Wu, Arora, Wang, Geiger, Jurafsky, Manning, and Potts]{wu2024reft}
Zhengxuan Wu, Aryaman Arora, Zheng Wang, Atticus Geiger, Dan Jurafsky, Christopher~D Manning, and Christopher Potts.
\newblock Reft: Representation finetuning for language models.
\newblock \emph{Advances in Neural Information Processing Systems}, 37:\penalty0 63908--63962, 2024{\natexlab{b}}.

\bibitem[Xiao et~al.(2025)Xiao, Yan, Hong, Cai, Jiang, Hu, Shen, Wang, and Snoek]{xiao2025dynaprompt}
Zehao Xiao, Shilin Yan, Jack Hong, Jiayin Cai, Xiaolong Jiang, Yao Hu, Jiayi Shen, Qi~Wang, and Cees~GM Snoek.
\newblock Dynaprompt: Dynamic test-time prompt tuning.
\newblock \emph{arXiv preprint arXiv:2501.16404}, 2025.

\bibitem[Yadav et~al.(2023)Yadav, Tam, Choshen, Raffel, and Bansal]{yadav2023ties}
Prateek Yadav, Derek Tam, Leshem Choshen, Colin~A Raffel, and Mohit Bansal.
\newblock Ties-merging: Resolving interference when merging models.
\newblock \emph{Advances in Neural Information Processing Systems}, 36:\penalty0 7093--7115, 2023.

\bibitem[Yang et~al.(2024)Yang, Yang, Zhang, Hui, Zheng, Yu, Li, Liu, Huang, Wei, et~al.]{yang2024qwen2}
An~Yang, Baosong Yang, Beichen Zhang, Binyuan Hui, Bo~Zheng, Bowen Yu, Chengyuan Li, Dayiheng Liu, Fei Huang, Haoran Wei, et~al.
\newblock Qwen2. 5 technical report.
\newblock \emph{arXiv preprint arXiv:2412.15115}, 2024.

\bibitem[Yu et~al.(2024)Yu, Yu, Yu, Huang, and Li]{yu2024language}
Le~Yu, Bowen Yu, Haiyang Yu, Fei Huang, and Yongbin Li.
\newblock Language models are super mario: Absorbing abilities from homologous models as a free lunch.
\newblock In \emph{Forty-first International Conference on Machine Learning}, 2024.

\bibitem[Yu et~al.(2020)Yu, Kumar, Gupta, Levine, Hausman, and Finn]{yu2020gradient}
Tianhe Yu, Saurabh Kumar, Abhishek Gupta, Sergey Levine, Karol Hausman, and Chelsea Finn.
\newblock Gradient surgery for multi-task learning.
\newblock \emph{Advances in neural information processing systems}, 33:\penalty0 5824--5836, 2020.

\bibitem[Zadouri et~al.(2023)Zadouri, {\"U}st{\"u}n, Ahmadian, Ermi{\c{s}}, Locatelli, and Hooker]{zadouri2023pushing}
Ted Zadouri, Ahmet {\"U}st{\"u}n, Arash Ahmadian, Beyza Ermi{\c{s}}, Acyr Locatelli, and Sara Hooker.
\newblock Pushing mixture of experts to the limit: Extremely parameter efficient moe for instruction tuning.
\newblock \emph{arXiv preprint arXiv:2309.05444}, 2023.

\bibitem[Zaken et~al.(2021)Zaken, Ravfogel, and Goldberg]{zaken2021bitfit}
Elad~Ben Zaken, Shauli Ravfogel, and Yoav Goldberg.
\newblock Bitfit: Simple parameter-efficient fine-tuning for transformer-based masked language-models.
\newblock \emph{arXiv preprint arXiv:2106.10199}, 2021.

\bibitem[Zeng et~al.(2019)Zeng, Chen, Cui, and Yu]{zeng2019continual}
Guanxiong Zeng, Yang Chen, Bo~Cui, and Shan Yu.
\newblock Continual learning of context-dependent processing in neural networks.
\newblock \emph{Nature Machine Intelligence}, 1\penalty0 (8):\penalty0 364--372, 2019.

\bibitem[Zeng et~al.(2025)Zeng, He, You, Hao, Tsai, Yamada, and Zhao]{zeng2025efficient}
Siqi Zeng, Yifei He, Weiqiu You, Yifan Hao, Yao-Hung~Hubert Tsai, Makoto Yamada, and Han Zhao.
\newblock Efficient model editing with task vector bases: A theoretical framework and scalable approach.
\newblock \emph{arXiv preprint arXiv:2502.01015}, 2025.

\bibitem[Zhang et~al.(2025)Zhang, You, Panda, and Goldstein]{zhang2025lori}
Juzheng Zhang, Jiacheng You, Ashwinee Panda, and Tom Goldstein.
\newblock Lori: Reducing cross-task interference in multi-task low-rank adaptation.
\newblock \emph{arXiv preprint arXiv:2504.07448}, 2025.

\bibitem[Zhang et~al.(2023{\natexlab{a}})Zhang, Zhang, Shi, Chu, and Li]{zhang2023lora}
Longteng Zhang, Lin Zhang, Shaohuai Shi, Xiaowen Chu, and Bo~Li.
\newblock Lora-fa: Memory-efficient low-rank adaptation for large language models fine-tuning.
\newblock \emph{arXiv preprint arXiv:2308.03303}, 2023{\natexlab{a}}.

\bibitem[Zhang et~al.(2023{\natexlab{b}})Zhang, Chen, Bukharin, Karampatziakis, He, Cheng, Chen, and Zhao]{zhang2023adalora}
Qingru Zhang, Minshuo Chen, Alexander Bukharin, Nikos Karampatziakis, Pengcheng He, Yu~Cheng, Weizhu Chen, and Tuo Zhao.
\newblock Adalora: Adaptive budget allocation for parameter-efficient fine-tuning.
\newblock \emph{arXiv preprint arXiv:2303.10512}, 2023{\natexlab{b}}.

\bibitem[Zheng et~al.(2025)Zheng, Liu, Li, Chen, Yu, Gao, Dang, Liu, Men, Yang, et~al.]{zheng2025group}
Chujie Zheng, Shixuan Liu, Mingze Li, Xiong-Hui Chen, Bowen Yu, Chang Gao, Kai Dang, Yuqiong Liu, Rui Men, An~Yang, et~al.
\newblock Group sequence policy optimization.
\newblock \emph{arXiv preprint arXiv:2507.18071}, 2025.

\bibitem[Zhu et~al.(2024)Zhu, Greenewald, Nadjahi, Borde, Gabrielsson, Choshen, Ghassemi, Yurochkin, and Solomon]{zhu2024asymmetry}
Jiacheng Zhu, Kristjan Greenewald, Kimia Nadjahi, Haitz Saez De~Ocariz Borde, Rickard~Br{\"u}el Gabrielsson, Leshem Choshen, Marzyeh Ghassemi, Mikhail Yurochkin, and Justin Solomon.
\newblock Asymmetry in low-rank adapters of foundation models.
\newblock \emph{arXiv preprint arXiv:2402.16842}, 2024.

\bibitem[Zhu et~al.(2025)Zhu, Zhang, Wang, Yan, Sun, and Ji]{zhu2025patchwise}
Yao Zhu, Yunjian Zhang, Zizhe Wang, Xiu Yan, Peng Sun, and Xiangyang Ji.
\newblock Patchwise cooperative game-based interpretability method for large vision-language models.
\newblock \emph{Transactions of the Association for Computational Linguistics}, 13:\penalty0 744--759, 2025.

\bibitem[Zou et~al.(2025{\natexlab{a}})Zou, Mao, Qu, Wang, and Ji]{zou2025utility}
Heming Zou, Yixiu Mao, Yun Qu, Qi~Wang, and Xiangyang Ji.
\newblock Utility-diversity aware online batch selection for llm supervised fine-tuning.
\newblock \emph{arXiv preprint arXiv:2510.16882}, 2025{\natexlab{a}}.

\bibitem[Zou et~al.(2025{\natexlab{b}})Zou, Zang, and Ji]{zou2025structural}
Heming Zou, Yunliang Zang, and Xiangyang Ji.
\newblock Structural features of the fly olfactory circuit mitigate the stability-plasticity dilemma in continual learning.
\newblock \emph{arXiv preprint arXiv:2502.01427}, 2025{\natexlab{b}}.

\bibitem[Zou et~al.(2025{\natexlab{c}})Zou, Zang, Xu, and Ji]{zou2025fly}
Heming Zou, Yunliang Zang, Wutong Xu, and Xiangyang Ji.
\newblock Fly-cl: A fly-inspired framework for enhancing efficient decorrelation and reduced training time in pre-trained model-based continual representation learning.
\newblock \emph{arXiv preprint arXiv:2510.16877}, 2025{\natexlab{c}}.

\end{thebibliography}

%%%%%%%%%%%%%%%%%%%%%%%%%%%%%%%%%%%%%%%%%%%%%%%%%%%%%%%%%%%%

\newpage
\section*{NeurIPS Paper Checklist}

\begin{enumerate}

\item {\bf Claims}
    \item[] Question: Do the main claims made in the abstract and introduction accurately reflect the paper's contributions and scope?
    \item[] Answer: \answerYes{} % Replace by \answerYes{}, \answerNo{}, or \answerNA{}.
    \item[] Justification: We have summarized the contribution and scope in the abstract, introduction, and conclusion. Especially, we list our contribution in the last paragraph of the introduction.

\item {\bf Limitations}
    \item[] Question: Does the paper discuss the limitations of the work performed by the authors?
    \item[] Answer: \answerYes{} % Replace by \answerYes{}, \answerNo{}, or \answerNA{}.
    \item[] Justification: We have discussed the possible limitations in Appendix \ref{App:limit_future}.

\item {\bf Theory assumptions and proofs}
    \item[] Question: For each theoretical result, does the paper provide the full set of assumptions and a complete (and correct) proof?
    \item[] Answer: \answerYes{} % Replace by \answerYes{}, \answerNo{}, or \answerNA{}.
    \item[] Justification: We have listed theorems in Section \ref{Sec:flylora} of the main text, with complete proofs presented in Appendix \ref{App:theory}.

    \item {\bf Experimental result reproducibility}
    \item[] Question: Does the paper fully disclose all the information needed to reproduce the main experimental results of the paper to the extent that it affects the main claims and/or conclusions of the paper (regardless of whether the code and data are provided or not)?
    \item[] Answer: \answerYes{} % Replace by \answerYes{}, \answerNo{}, or \answerNA{}.
    \item[] Justification: We have listed the training details in Appendix \ref{app:experimental_setting}, from which people can reproduce the main experimental results based on it. We also provide code in the supplementary materials.

\item {\bf Open access to data and code}
    \item[] Question: Does the paper provide open access to the data and code, with sufficient instructions to faithfully reproduce the main experimental results, as described in supplemental material?
    \item[] Answer: \answerYes{} % Replace by \answerYes{}, \answerNo{}, or \answerNA{}.
    \item[] Justification: We provide code in the supplementary materials. The open-source datasets and models are listed in Appendix \ref{app:experimental_setting}.

\item {\bf Experimental setting/details}
    \item[] Question: Does the paper specify all the training and test details (e.g., data splits, hyperparameters, how they were chosen, type of optimizer, etc.) necessary to understand the results?
    \item[] Answer: \answerYes{} % Replace by \answerYes{}, \answerNo{}, or \answerNA{}.
    \item[] Justification: We discuss the experimental setup and training details in Section \ref{Sec:experiment} and Appendix \ref{app:experimental_setting}.

\item {\bf Experiment statistical significance}
    \item[] Question: Does the paper report error bars suitably and correctly defined or other appropriate information about the statistical significance of the experiments?
    \item[] Answer: \answerYes{} % Replace by \answerYes{}, \answerNo{}, or \answerNA{}.
    \item[] Justification: We report the error bar over three random seeds.

\item {\bf Experiments compute resources}
    \item[] Question: For each experiment, does the paper provide sufficient information on the computer resources (type of compute workers, memory, time of execution) needed to reproduce the experiments?
    \item[] Answer: \answerYes{} % Replace by \answerYes{}, \answerNo{}, or \answerNA{}.
    \item[] Justification: We summarize our computational resources in Appendix \ref{app:experimental_setting}.
    
\item {\bf Code of ethics}
    \item[] Question: Does the research conducted in the paper conform, in every respect, with the NeurIPS Code of Ethics \url{https://neurips.cc/public/EthicsGuidelines}?
    \item[] Answer: \answerYes{} % Replace by \answerYes{}, \answerNo{}, or \answerNA{}.
    \item[] Justification: We have read the Code of Ethics and make sure to preserve anonymity.

\item {\bf Broader impacts}
    \item[] Question: Does the paper discuss both potential positive societal impacts and negative societal impacts of the work performed?
    \item[] Answer: \answerYes{} % Replace by \answerYes{}, \answerNo{}, or \answerNA{}.
    \item[] Justification: We have discuss potential societal impacts in Appendix \ref{App:broader_impact}.
    
\item {\bf Safeguards}
    \item[] Question: Does the paper describe safeguards that have been put in place for responsible release of data or models that have a high risk for misuse (e.g., pretrained language models, image generators, or scraped datasets)?
    \item[] Answer: \answerNA{} % Replace by \answerYes{}, \answerNo{}, or \answerNA{}.
    \item[] Justification: Our work poses no such risks.

\item {\bf Licenses for existing assets}
    \item[] Question: Are the creators or original owners of assets (e.g., code, data, models), used in the paper, properly credited and are the license and terms of use explicitly mentioned and properly respected?
    \item[] Answer: \answerYes{} % Replace by \answerYes{}, \answerNo{}, or \answerNA{}.
    \item[] Justification: We have used open-source code and datasets, and have appropriately cited them.

\item {\bf New assets}
    \item[] Question: Are new assets introduced in the paper well documented and is the documentation provided alongside the assets?
    \item[] Answer: \answerYes{} % Replace by \answerYes{}, \answerNo{}, or \answerNA{}.
    \item[] Justification: Our paper does not release new assets.

\item {\bf Crowdsourcing and research with human subjects}
    \item[] Question: For crowdsourcing experiments and research with human subjects, does the paper include the full text of instructions given to participants and screenshots, if applicable, as well as details about compensation (if any)? 
    \item[] Answer: \answerNA{} % Replace by \answerYes{}, \answerNo{}, or \answerNA{}.
    \item[] Justification: Our paper does not involve crowdsourcing nor research with human subjects.

\item {\bf Institutional review board (IRB) approvals or equivalent for research with human subjects}
    \item[] Question: Does the paper describe potential risks incurred by study participants, whether such risks were disclosed to the subjects, and whether Institutional Review Board (IRB) approvals (or an equivalent approval/review based on the requirements of your country or institution) were obtained?
    \item[] Answer: \answerNA{} % Replace by \answerYes{}, \answerNo{}, or \answerNA{}.
    \item[] Justification: Our paper does not involve crowdsourcing nor research with human subjects.

\item {\bf Declaration of LLM usage}
    \item[] Question: Does the paper describe the usage of LLMs if it is an important, original, or non-standard component of the core methods in this research? Note that if the LLM is used only for writing, editing, or formatting purposes and does not impact the core methodology, scientific rigorousness, or originality of the research, declaration is not required.
    %this research? 
    \item[] Answer: \answerNA{} % Replace by \answerYes{}, \answerNo{}, or \answerNA{}.
    \item[] Justification: The core method development in this research does not involve LLMs as any important, original, or non-standard components.

\end{enumerate}

\newpage
\appendix

\section{Theoretical Analysis} \label{App:theory}

\subsection{Distance Preserving Property for Sparse Random Projection} \label{App:distance_preserving}
In this section, we prove that the fixed sparse random projection matrix $\bm{A}$ satisfies the distance-preserving property. This demonstrates that a fixed projection $\bm{A}$ can function like a hash router without an explicit router. Our result extends the well-known Johnson-Lindenstrauss Lemma \cite{johnson1984extensions}. First, \cite[Corollary 3.5]{bruhin2022bioinspired} provides the following bound:
\begin{theorem} \label{thm:initial_bound}
    Let $\bm{A}\in\mathbb{R}^{r \times n}$ be a random matrix whose entries $\bm{A}_{ij}$ are sampled independently and randomly from a distribution that is symmetric around the origin with $\mathbb{E}(\bm{A}_{ij}^2)=\sigma^2>0$.
    \begin{enumerate}
        \item Suppose $C = \mathbb{E}(\bm{A}_{ij}^4) < \infty$. Then, for any $\epsilon > 0$,
        $$\mathbb{P}\left(\| \frac{1}{\sqrt{r}} \bm{A}\bm{x}\|^2 \leq \sigma^2(1- \epsilon)\|\bm{x}\|^2\right) \leq \exp\left(- \frac{(\epsilon^2 - \epsilon^3)r}{2(\frac{1}{\sigma^4} C+1)}\right), \quad\text{for all } \bm{x} \in \mathbb{R}^n.$$
        \item Suppose $\exists L > 0$ such that for any integer $k > 0$, $\mathbb{E}(\bm{A}_{ij}^{2k}) \leq \sigma^{2k} \frac{(2k)!}{2^k k!} L^{2k}$. Then, for any $\epsilon > 0$,
        $$\mathbb{P}\left(\|\frac{1}{\sqrt{r}} \bm{A}\bm{x}\|^2 \geq \sigma^2 (1+\epsilon)L^2\|\bm{x}\|^2\right) \leq \exp\left(-(\epsilon^2 - \epsilon^3)\frac{r}{4}\right), \quad\text{for all } \bm{x} \in \mathbb{R}^n.$$
    \end{enumerate}
\end{theorem}
According to the definition of $\bm{A}$ mentioned in Section 3.1, the second moment of $\bm{A}_{ij}$ satisfies $\mathbb{E}(\bm{A}_{ij}^2)=\frac{p}{nr^2}>0$. So, the $2k$-th moment of $\bm{A}_{ij}$ is given by $\mathbb{E}[\bm{A}_{ij}^{2k}]=\frac{p}{n}\cdot\mathbb{E}[\bm{X}_{ij}^{2k}]+\left(1-\frac{p}{n}\right)\cdot0=\frac{p}{n}\cdot\mathbb{E}[\bm{X}_{ij}^{2k}]$, where $\bm{X}_{ij}\sim\mathcal{N}(0,\frac{1}{r^2})$. From the property of Gaussian distribution, we derive $\mathbb{E}[\bm{X}_{ij}^{2k}]=(2k-1)!!\cdot\left(\frac{1}{r^2}\right)^{2k}=\frac{(2k)!}{2^kk!}\left(\frac{1}{r^2}\right)^{2k}$, where $!!$ denotes the double factorial. This leads to $\mathbb{E}[\bm{A}_{ij}^4]=\frac{3p}{nr^4}$, which is clearly finite. To satisfy the inequality $\mathbb{E}[\bm{A}_{ij}^{2k}]=\frac{(2k)!}{2^kk!}\sigma^{2k}\le\sigma^{2k} \frac{(2k)!}{2^k k!} L^{2k}$, we require $L\ge \log_{2k}\frac{p}{r}$. Due to monotonicity, simply choosing $L=\log_2\frac{p}{r}$ always satisfies the condition specified in Theorem \ref{thm:initial_bound}. Combining both bounds from Theorem \ref{thm:initial_bound}, we summarize the final result in Theorem \ref{thm:distance}.
\begin{equation}
\begin{split}
    &\mathbb{P}\left( (1-\epsilon) \|\bm{x}-\bm{y}\|^2 \leq \frac{1}{r\sigma^2}\|\bm{A}\bm{x}-\bm{A}\bm{y}\|^2 \leq (1+\epsilon) \|\bm{x}-\bm{y}\|^2\right) = \\& 1- 
    \mathbb{P}\left(\frac{1}{r}\|\bm{A}\bm{x}-\bm{A}\bm{y}\|^2 < (1-\epsilon)\sigma^2 \|\bm{x}-\bm{y}\|^2 \right) - \\&
    \mathbb{P}\left( \frac{1}{r}\|\bm{A}\bm{x}-\bm{A}\bm{y}\|^2 > (1+\epsilon) \sigma^2 \|\bm{x}-\bm{y}\|^2\right),
\end{split}
\end{equation}
\begin{theorem} \label{thm:distance}
    Given the matrix $\bm{A}\in\mathbb{R}^{r\times n}$  with each entry i.i.d. from $\mathcal{N}(0, \frac{1}{r^2})$, and set to $0$ otherwise, for any $\epsilon>0$,
    $$\mathbb{P}\left( (1-\epsilon) \|\bm{x}-\bm{y}\|^2 \leq \frac{1}{r\sigma^2}\|\bm{A}\bm{x}- \bm{A}\bm{y}\|^2 \leq (1+ \epsilon)  \|\bm{x}-\bm{y}\|^2\right) \geq 1-e^{-(\epsilon^2 - \epsilon^3) \frac{r}{4}} - e^{- \frac{(\epsilon^2 - \epsilon^3)r}{2(\frac{3p}{n} + 1)}},$$
    for any input embeddings $\bm{x},\bm{y}\in\mathbb{R}^n$, where $\sigma^2=\frac{p}{nr^2}$.
\end{theorem}
Since this construction does not differ significantly from our desired construction for $\bm{A}$, which includes an additional constraint on non-zero entries per row and is more consistent with biological observations in the fly olfactory circuit, we use this more analytically tractable form as a surrogate to study the distance-preserving property. In practice, these two construction methods show negligible performance differences.

\subsection{Top-k Activation Promotes Rank-Wise Decoupling} \label{App:topk_decoupling}
Let $\bm{\Lambda}\in\mathrm{diag}(\lambda_1,\dots,\lambda_r)\in\mathbb{R}^{r\times r}$ denote a binary mask where $\lambda_i=1$ if column $i$ is activated by top-$k$. The relation of the masked gradient between top-$k$ version ($\frac{\partial \mathcal{L}}{\partial \tilde{\bm{B}}}$) and the dense version ($\frac{\partial \mathcal{L}}{\partial \bm{B}}$) is:
\begin{equation}
    \frac{\partial \mathcal{L}}{\partial \tilde{\bm{B}}} = \frac{\partial \mathcal{L}}{\partial \bm{B}} \bm{\Lambda} \in \mathbb{R}^{m \times r}.
\end{equation}
Let $\tilde{\bm{g}}_i$ and $\bm{g}_i\in\mathbb{R}^m$ denote the gradient vectors for the $i$-th column of $\frac{\partial \mathcal{L}}{\partial \tilde{\bm{B}}}$ and $\frac{\partial \mathcal{L}}{\partial \bm{B}}$, respectively. Their corresponding cross-column covariance are:
\begin{align}
    \bm{\Sigma}_{(i,j)} &= \mathbb{E}\left[\bm{g}_i^\top\bm{g}_j\right]-\mathbb{E}\left[\bm{g}_i\right]^\top\mathbb{E}\left[\bm{g}_j\right], \\
    \tilde{\bm{\Sigma}}_{(i,j)} &= \mathbb{E}\left[\tilde{\bm{g}}_i^\top\tilde{\bm{g}}_j\right]-\mathbb{E}\left[\tilde{\bm{g}}_i\right]^\top\mathbb{E}\left[\tilde{\bm{g}}_j\right],
\end{align}
We can assume $\mathbb{E}[\tilde{\bm{g}}_i]=\mathbb{E}[\tilde{\bm{g}}_j]=\mathbb{E}[\bm{g}_i]=\mathbb{E}[\bm{g}_j]=\bm{0}_m$, these simplify to:
\begin{align}
    \bm{\Sigma}_{(i,j)} &=\mathbb{E}\left[\bm{g}_i^\top\bm{g}_j\right], \\
    \tilde{\bm{\Sigma}}_{(i,j)} &= \mathbb{E}\left[\tilde{\bm{g}}_i^\top\tilde{\bm{g}}_j\right].
\end{align}
Thus, the expected covariance depends only on the expected inner product of gradient vectors. From Assumption \ref{asm:Uniform_Sparse_Activation}, the difference between $\tilde{\bm{\Sigma}}_{(i,j)}$, $\bm{\Sigma}_{(i,j)}$ is factored by the co-activation probability of columns $i$ and $j$:
\begin{equation}
    \mathbb{P}(\lambda_i=1\cap \lambda_j=1)=\frac{\binom{r-2}{k-2}}{\binom{r}{k}}=\frac{k(k-1)}{r(r-1)}\approx\frac{k^2}{r^2}.
\end{equation}
This leads to the following theorem:
\begin{theorem}[Covariance Reduction Under top-$k$]
Let $\tilde{\bm{\Sigma}}$ and $\bm{\Sigma}$ denote the gradient covariance matrices with and without top-$k$ activation. When $r > k$, the off-diagonal entries scale as:
\begin{equation}
\mathbb{E}[\tilde{\bm{\Sigma}}_{(i,j)}] \approx \mathbb{E}[\bm{\Sigma}_{(i,j)}] \cdot \frac{k^2}{r^2}, \quad \forall i \neq j.
\end{equation}
\end{theorem}

\subsection{Random Projection Induces Approximate Subspace Orthogonality} \label{App:subspace_orthogonality}
The orthogonality properties of random projections form the theoretical foundation for FlyLoRA's effectiveness in model merging. Consider two independent random projection matrices $\bm{A}_i,\bm{A}_j\in\mathbb{R}^{r\times n}$ with entries distributed as specified in Section \ref{sec:formulation}. The expectation of their product reveals:
\begin{equation}
    \mathbb{E}[\bm{A}_i\bm{A}_j^\top]=\mathbb{E}\left[\sum_{k=1}^n(\bm{A}_i)_{mk}(\bm{A}_j)_{lk}\right]=\sum_{k=1}^n\mathbb{E}[(\bm{A}_i)_{mk}]\mathbb{E}[(\bm{A}_j)_{lk}]=\bm{0}_{r\times r}
\end{equation}
This zero-expectation result follows from the independence and zero-mean property of the random matrices. To quantify how tightly the product concentrates around zero, we analyze its variance:
\begin{align}
\mathrm{Var}\left((\bm{A}_i\bm{A}_j^T)_{ml}\right) 
    &= \sum_{k=1}^n \Big[\mathrm{Var}\big((\bm{A}_i)_{mk}\big)\mathrm{Var}\big((\bm{A}_j)_{lk}\big) \notag + \mathrm{Var}\big((\bm{A}_i)_{mk}\big)\big(\mathbb{E}(\bm{A}_j)_{lk}\big)^2 \notag \\
    &\quad + \mathrm{Var}\big((\bm{A}_j)_{lk}\big)\big(\mathbb{E}(\bm{A}_i)_{mk}\big)^2 \Big] \notag \\
    &= \sum_{k=1}^n \mathrm{Var}\big((\bm{A}_i)_{mk}\big)\mathrm{Var}\big((\bm{A}_j)_{lk}\big) \notag \\
    &= n\sigma^4 = \frac{p^2}{nr^4}
\end{align}
Applying Chebyshev's inequality, we bound the probability of large deviations for each entry:
\begin{equation}
    \mathbb{P}\left(|(\bm{A}_i\bm{A}_j^\top)_{ml}|\geq \epsilon\right)\leq\frac{\mathrm{Var}\left((\bm{A}_i\bm{A}_j^\top)_{ml}\right)}{\epsilon^2}=\frac{p^2}{nr^4\epsilon^2}
\end{equation}
The Frobenius norm characterization gives us:$\|\bm{A}_i\bm{A}_j^\top\|_F^2=\sum_{m=1}^r\sum_{l=1}^r(\bm{A}_i\bm{A}_j^\top)_{ml}^2.$ To analyze its probabilistic behavior, we employ the union bound principle: for any events $E_{ij}$ defined as $|(\bm{A}_i\bm{A}_j^\top)_{ml}|\geq\epsilon$, we have $\mathbb{P}\left(\bigcup_{m,l}E_{ml}\right)\leq\sum_{m,l}\mathbb{P}(E_{ml})$. Using this union bound over all $r^2$ entries yields:
\begin{equation}
    \mathbb{P}\left(\|\bm{A}_i\bm{A}_j^\top\|_F\geq \epsilon r\right)\leq\sum_{m=1}^r\sum_{l=1}^r\mathbb{P}\left(|(\bm{A}_i\bm{A}_j^\top)_{ml}|\geq \epsilon\right)\leq r^2 \cdot \frac{p^2}{n r^4 \epsilon^2} = \frac{p^2}{n r^2 \epsilon^2}.
\end{equation}
Since the spectral norm is bounded by the Frobenius norm ($\|\bm{A}_i^\top \bm{A}_j\|_2\leq\|\bm{A}_i^\top \bm{A}_j\|_F$), we conclude that when $n\gg\frac{p^2}{r^2\epsilon^2}$, the subspaces spanned by $\bm{A}_i$ and $\bm{A}_j$ are approximately orthogonal with high probability.
\begin{theorem}[Approximate Subspace Orthogonality]
    For independent random matrices $\bm{A}_i,\bm{A}_j\in \mathbb{R}^{r\times n}$ with sparse Gaussian entries ($\mathcal{N}(0, \frac{1}{r^2})$ for $p<n$ randomly selected entries per row), the following holds:
    \begin{enumerate}
    \item\textbf{Exact mean orthogonality}: $\mathbb{E}[\bm{A}_i \bm{A}_j^\top]=\bm{0}_{r\times r}$
    \item\textbf{Polynomially decaying correlations}: $\mathbb{P}(\| \bm{A}_i\bm{A}_j^\top\|_2\ge \epsilon r)\le \frac{p^2}{nr^2\epsilon^2}$
    \end{enumerate}
\end{theorem}
We demonstrate the approximate orthogonality between distinct LoRA components $\bm{B}_i\bm{A}_i$ and $\bm{B}_j\bm{A}_j$ through Frobenius inner product analysis following \cite{tang2023parameter}:
\begin{align}
    \langle\bm{B}_j\bm{A}_j, \bm{B}_i\bm{A}_i\rangle_F &= \mathrm{tr}\left(\left(\bm{B}_j\bm{A}_j\right)^\top\left(\bm{B}_i\bm{A}_i\right)\right) \notag \\
    &= \mathrm{tr}\left(\bm{A}_j^\top\bm{B}_j^\top\bm{B}_i\bm{A}_i\right) \notag \\
    &= \mathrm{tr}\left(\bm{B}_j^\top\bm{B}_i\bm{A}_i\bm{A}_j^\top\right) \notag \\
    &\approx \mathrm{tr}\left(\bm{B}_j^\top\bm{B}_i\cdot\bm{0}_{r \times r}\right) \notag \\
    &\approx 0
\end{align}
These analysis demonstrates that random projections naturally create nearly orthogonal subspaces, which helps prevent interference between different experts in FlyLoRA. The small residual correlation becomes negligible when input dimension $n\rightarrow\infty$, leading to the effectiveness of our sparse random projection approach for model merging.

Formally, considering the conventional LoRA merging scheme:
\begin{equation}
    \bm{W}'=\bm{W}_0 + \sum_{i=1}^t w_i\bm{B}_i\bm{A}_i
\end{equation}
When multiple FlyLoRA modules ($\bm{B}_i\bm{A}_i$) are approximately orthogonal, the squared Frobenius norm of the merged weight matrix can be decomposed into the weighted sum of individual module norms.
\begin{align}
    \left\|\sum_{i=1}^tw_i\bm{B}_i\bm{A}_i\right\|_F^2 &= \sum_{i=1}^tw_i^2\|\bm{B}_i\bm{A}_i\|_F^2+\sum_{i\neq j}w_iw_j\langle\bm{B}_i\bm{A}_i,\bm{B}_j\bm{A}_j\rangle_F \\
    &\approx \sum_{i=1}^tw_i^2\|\bm{B}_i\bm{A}_i\|_F^2,
\end{align}
which aligns with the ``Weight disentanglement'' property for task arithmetic \cite{ortiz2023task}. We summarize these findings in the following corollary:
\begin{corollary}
    Let $\bm{A}_i,\bm{A}_j \in \mathbb{R}^{r \times n}$ be fixed sparse random projections after initialization. Then for any learned matrices $\bm{B}_i\bm{A}_i$ and $\bm{B}_j\bm{A}_j$ the following properties hold:
    \begin{enumerate}
        \item \textbf{Pairwise Orthogonality}:
        $$\langle \bm{B}_i\bm{A}_i, \bm{B}_j\bm{A}_j \rangle_F \approx 0 \quad \text{for} \quad i \neq j$$
        \item \textbf{Orthogonality's Outcome in Merging}:
        $$\left\|\sum_{i=1}^t w_i\bm{B}_i\bm{A}_i\right\|_F^2 \approx \sum_{i=1}^t w_i^2\|\bm{B}_i\bm{A}_i\|_F^2$$
    \end{enumerate}
\end{corollary}
Through sparse random projections, FlyLoRA inherently constructs nearly orthogonal subspaces for parameter merging from different tasks.

\section{Additional Results} \label{App:additional_results}

\subsection{Evaluation on Larger Models}
We further conducted experiments using the Qwen-2.5-14B model, as shown in Tables \ref{tab:single_larger_model} and \ref{tab:multi_larger_model}, following the settings of Tables \ref{tab:single_results} and \ref{tab:multi_results}. The results show that FlyLoRA remains superior in accuracy (for both single-task and multi-task settings) and is more parameter-efficient. We encountered no memory or convergence bottlenecks when training FlyLoRA on the 14B model, which consistently outperforms LoRA and Split-LoRA. This scalability confirms that our method's benefits extend effectively to larger model architectures without compromising training stability.
\begin{table}[h]
    \centering
    \caption{
        \textbf{Performance Comparison of LoRA Variants in Single-task Evaluation using Qwen-2.5-14B.} We evaluate various methods across four benchmarks: MMLU, ScienceQA, GSM8K (accuracy), and HumanEval (Pass@k), with all metrics reported in percentage (\%). Param(\%) indicates the percentage of activated trainable parameters relative to Full FT. The best results are highlighted in \textbf{bold}.
    }
    \vspace{+2pt}
    \resizebox{0.7\textwidth}{!}{
    \setlength{\tabcolsep}{1mm}{
    \begin{tabular}{cccccc}
        \toprule
        \textbf{Method} & Param(\%) & \textbf{MMLU} & \textbf{ScienceQA} & \textbf{GSM8K} & \textbf{HumanEval} \\
        \midrule
        LoRA$_{(r=8)}$ & 0.23 & 56.74\scriptsize{$\pm$0.56} & 95.62\scriptsize{$\pm$0.18} & 83.08\scriptsize{$\pm$0.74} & 51.69\scriptsize{$\pm$1.48} \\
        LoRA$_{(r=32)}$ & 0.93 & 59.35\scriptsize{$\pm$0.79} & 97.05\scriptsize{$\pm$0.22} & 85.31\scriptsize{$\pm$0.25} & 54.80\scriptsize{$\pm$0.76} \\
        Split-LoRA$_{(4\times8)}$ & 0.29 & 58.26\scriptsize{$\pm$1.13} & 96.85\scriptsize{$\pm$0.35} & 84.88\scriptsize{$\pm$0.51} & 54.65\scriptsize{$\pm$0.59} \\
        \midrule
        \cellcolor{gray!20}FlyLoRA & \cellcolor{gray!20}\textbf{0.12} & \cellcolor{gray!20}\textbf{60.17\scriptsize{$\pm$1.08}} & \cellcolor{gray!20}\textbf{97.37\scriptsize{$\pm$0.32}} & \cellcolor{gray!20}\textbf{85.96\scriptsize{$\pm$0.89}} & \cellcolor{gray!20}\textbf{56.42\scriptsize{$\pm$1.16}} \\
        \bottomrule
    \end{tabular}}}
    \label{tab:single_larger_model}
    \vspace{-10pt}
\end{table}
\begin{table}[h]
    \centering
    \caption{
        \textbf{Multi-task Performance Comparison using Qwen-2.5-14B.} We evaluate LoRA variants across MMLU, ScienceQA, GSM8K (accuracy), and HumanEval (Pass@k) benchmarks. The table shows the relative performance drop ($\Delta$\%) before and after merging. The best results are highlighted in \textbf{bold}.
    }
    \vspace{+2pt}
    \resizebox{0.6\textwidth}{!}{
    \setlength{\tabcolsep}{1mm}{
    \begin{tabular}{ccccc}
        \toprule
        \textbf{Method} & \textbf{MMLU} & \textbf{ScienceQA} & \textbf{GSM8K} & \textbf{HumanEval} \\
        \midrule
        LoRA$_{(r=8)}$ & -13.75 & -25.20 & -11.43 & -18.60 \\
        LoRA$_{(r=32)}$ & -8.91 & -20.45 & -7.62 & -16.34 \\
        Split-LoRA$_{(4\times8)}$ & -7.48 & -21.97 & -6.05 & -14.87 \\
        \midrule
        \cellcolor{gray!20}FlyLoRA & \cellcolor{gray!20}\textbf{-4.35} & \cellcolor{gray!20}\textbf{-17.89} & \cellcolor{gray!20}\textbf{-2.18} & \cellcolor{gray!20}\textbf{-11.72} \\
        \bottomrule
    \end{tabular}}}
    \label{tab:multi_larger_model}
    \vspace{-10pt}
\end{table}

\subsection{More Baseline into Comparison}
We further compare several strong and widely used baselines---AdaLoRA \cite{zhang2023adalora} (adaptive rank allocation), SoRA \cite{ding2023sparse} (sparse adaptation), and HydraLoRA \cite{tian2024hydralora} (MoE-based)---using Qwen-2.5-7B, as shown in Tables \ref{tab:single_more_baseline} and \ref{tab:multi_more_baseline}. The results demonstrate that FlyLoRA remains superior in terms of accuracy and efficiency in both single-task learning and multi-task merging scenarios. Notably, FlyLoRA achieves this performance while maintaining a simpler training pipeline, as it avoids the additional hyperparameter tuning required by adaptive and sparse methods.
\begin{table}[h]
    \centering
    \caption{
        \textbf{Performance Comparison with More LoRA Variants in Single-task Evaluation using Qwen-2.5-7B.} We evaluate various methods across four benchmarks: MMLU, ScienceQA, GSM8K (accuracy), and HumanEval (Pass@k), with all metrics reported in percentage (\%). Param(\%) indicates the percentage of activated trainable parameters relative to Full FT. The best results are highlighted in \textbf{bold}.
    }
    \vspace{+2pt}
    \resizebox{0.8\textwidth}{!}{
    \setlength{\tabcolsep}{1mm}{
    \begin{tabular}{cccccc}
        \toprule
        \textbf{Method} & Param(\%) & \textbf{MMLU} & \textbf{ScienceQA} & \textbf{GSM8K} & \textbf{HumanEval} \\
        \midrule
        AdaLoRA$_{(r=8)}$ & 0.26 & 51.22\scriptsize{$\pm$0.21} & 93.48\scriptsize{$\pm$0.28} & 77.65\scriptsize{$\pm$0.14} & 47.96\scriptsize{$\pm$1.34} \\
        SoRA$_{(r=8)}$ & 0.19 & 50.89\scriptsize{$\pm$0.42} & 93.25\scriptsize{$\pm$0.20} & 78.46\scriptsize{$\pm$0.82} & 47.83\scriptsize{$\pm$0.94} \\
        HydraLoRA$_{(r=8,A=1,B=3)}$ & 0.52 & 53.05\scriptsize{$\pm$0.16} & 94.69\scriptsize{$\pm$0.34} & 79.31\scriptsize{$\pm$0.49} & 52.98\scriptsize{$\pm$1.57} \\
        \midrule
        \cellcolor{gray!20}FlyLoRA$_{(k=8)}$ & \cellcolor{gray!20}\textbf{0.13} & \cellcolor{gray!20}\textbf{53.68\scriptsize{$\pm$0.47}} & \cellcolor{gray!20}\textbf{95.55\scriptsize{$\pm$0.18}} & \cellcolor{gray!20}\textbf{80.82\scriptsize{$\pm$0.56}} & \cellcolor{gray!20}\textbf{54.34\scriptsize{$\pm$2.13}} \\
        \bottomrule
    \end{tabular}}}
    \label{tab:single_more_baseline}
    \vspace{-10pt}
\end{table}
\begin{table}[h]
    \centering
    \caption{
        \textbf{Multi-task Performance Comparison with More Baselines using Qwen-2.5-7B.} We evaluate LoRA variants across MMLU, ScienceQA, GSM8K (accuracy), and HumanEval (Pass@k) benchmarks. The table shows the relative performance drop ($\Delta$\%) before and after merging. The best results are highlighted in \textbf{bold}.
    }
    \vspace{+2pt}
    \resizebox{0.65\textwidth}{!}{
    \setlength{\tabcolsep}{1mm}{
    \begin{tabular}{ccccc}
        \toprule
        \textbf{Method} & \textbf{MMLU} & \textbf{ScienceQA} & \textbf{GSM8K} & \textbf{HumanEval} \\
        \midrule
        AdaLoRA$_{(r=8)}$ & -10.90 & -33.15 & +4.52 & -25.46 \\
        SoRA$_{(r=8)}$ & -9.45 & -34.81 & +4.05 & -26.50 \\
        HydraLoRA$_{(r=8,A=1,B=3)}$ & -6.22 & -34.67 & +4.21 & -24.94 \\
        \midrule
        \cellcolor{gray!20}FlyLoRA$_{(k=8)}$ & \cellcolor{gray!20}\textbf{+6.55} & \cellcolor{gray!20}\textbf{-23.77} & \cellcolor{gray!20}\textbf{+4.80} & \cellcolor{gray!20}\textbf{-21.23} \\
        \bottomrule
    \end{tabular}}}
    \label{tab:multi_more_baseline}
    \vspace{-10pt}
\end{table}

\subsection{Training Time and Memory Consumption}
We further conduct experiments comparing the training time and memory consumption of the LoRA variants discussed in Section \ref{Sec:experiment}. The results, presented in Table \ref{tab:time_memory}, demonstrate that LoRA$_{(r=32)}$ requires more training time and memory than LoRA$_{(r=8)}$ due to its higher rank. Split-LoRA$_{(4\times8)}$ shows intermediate values between these two approaches. Notably, FlyLoRA achieves both the fastest training times (resulting from having the fewest activated parameters) and the lowest memory consumption (mainly attributed to its frozen matrix $\bm{A}$ that significantly reduces memory for activation values \cite{zhang2023lora}). We also illustrate the theoretical memory consumption for these LoRA variants in Table \ref{tab:theoretical_memory}, which tightly aligns with the experimental results. It's clear to see that under a fixed total rank $r$, a larger number of experts $N$ causes Split-LoRA to require significantly more activated trainable parameters and memory consumption, degrading the efficiency of MoE-based LoRA methods. These results and analysis demonstrate the robustness of FlyLoRA's parameter efficiency compared to MoE-based LoRA methods.
\begin{table}[h]
    \centering
    \caption{
        \textbf{Training Time (Hours) and Memory Usage (GB) of LoRA Variants on Different Datasets and Architectures.} 
        Comparison of LoRA$_{(r=8)}$, LoRA$_{(r=32)}$, Split-LoRA$_{(4 \times 8)}$, and FlyLoRA$_{(k=8)}$ 
        fine-tuning Llama-3.1-8B and Qwen-2.5-7B on MMLU, ScienceQA, GSM8K, and CodeAlpaca-20k. The best results are highlighted in \textbf{bold}.
    }
    \vspace{+2pt}
    \resizebox{\textwidth}{!}{
    \begin{tabular}{lcccccccc}
        \toprule
        \multirow{2}{*}{\textbf{Metric}} & \multicolumn{4}{c}{\textbf{Llama-3.1-8B}} & \multicolumn{4}{c}{\textbf{Qwen-2.5-7B}} \\
        \cmidrule(lr){2-5} \cmidrule(lr){6-9}
        & \textbf{MMLU} & \textbf{ScienceQA} & \textbf{GSM8K} & \textbf{CodeAlpaca} & \textbf{MMLU} & \textbf{ScienceQA} & \textbf{GSM8K} & \textbf{CodeAlpaca} \\
        \midrule
        LoRA$_{(r=8)}$ \\
        \quad Training Time & 4.79h & 5.30h & 0.52h & 1.85h & 4.45h & 5.07h & 0.54h & 1.75h \\
        \quad Memory Usage & 12.5GB & 14.8GB & 20.1GB & 20.2GB & 14.7GB & 16.8GB & 22.9GB & 22.9GB \\
        \midrule
        LoRA$_{(r=32)}$ \\
        \quad Training Time & 5.09h & 5.61h & 0.58h & 1.95h & 4.76h & 5.39h & 0.60h & 1.87h \\
        \quad Memory Usage & 13.2GB & 15.3GB & 20.7GB & 20.7GB & 15.4GB & 17.3GB & 23.4GB & 23.4GB \\
        \midrule
        Split-LoRA$_{(4 \times 8)}$ \\
        \quad Training Time & 4.94h & 5.46h & 0.56h & 1.92h & 4.60h & 5.23h & 0.57h & 1.79h \\
        \quad Memory Usage & 12.8GB & 15.0GB & 20.4GB & 20.4GB & 15.0GB & 17.1GB & 23.2GB & 23.2GB \\
        \midrule
        FlyLoRA$_{(k=8)}$ \\
        \quad Training Time & \cellcolor{gray!20}\textbf{4.73h} & \cellcolor{gray!20}\textbf{5.23h} & \cellcolor{gray!20}\textbf{0.51h} & \cellcolor{gray!20}\textbf{1.82h} & \cellcolor{gray!20}\textbf{4.39h} & \cellcolor{gray!20}\textbf{4.99h} & \cellcolor{gray!20}\textbf{0.52h} & \cellcolor{gray!20}\textbf{1.70h} \\
        \quad Memory Usage & \cellcolor{gray!20}\textbf{10.6GB} & \cellcolor{gray!20}\textbf{10.7GB} & \cellcolor{gray!20}\textbf{10.9GB} & \cellcolor{gray!20}\textbf{10.9GB} & \cellcolor{gray!20}\textbf{12.1GB} & \cellcolor{gray!20}\textbf{12.2GB} & \cellcolor{gray!20}\textbf{12.4GB} & \cellcolor{gray!20}\textbf{12.4GB} \\
        \bottomrule
    \end{tabular}}
    \label{tab:time_memory}
    \vspace{-10pt}
\end{table}
\begin{table}[h]
    \centering
    \caption{
        \textbf{Theoretical Memory Consumption Comparison of Different LoRA Variants for a Single Linear Layer.} Param indicates the number of activated trainable parameters. Variables $d$, $r$, $k$, $b$, $s$, and $N$ represent hidden dimension, total rank, activation rank, batch size, sequence length, and number of experts, respectively. We record memory usage for weights, gradients, optimizer states, and activations in bytes. These results are calculated under 16-bit mixed-precision training settings.
    }
    \vspace{+2pt}
    \resizebox{0.9\textwidth}{!}{
    \setlength{\tabcolsep}{1mm}{
    \begin{tabular}{cccccc}
        \toprule
        \textbf{Method} & \textbf{Param} & \textbf{Weight} & \textbf{Gradient} & \textbf{Optimizer} & \textbf{Activation} \\
        \midrule
        LoRA & $2dr$ & $2(d^2 + 2dr)$ & $4dr$ & $24dr$ & $2bsd + 2bsr$ \\
        Split-LoRA & $2dk+dN$ & $2(d^2 + 2dr + dN)$ & $4dk + 2dN$ & $24dk + 12dN$ & $2bsd + 2bsk + 2bsN$ \\
        FlyLoRA & $dk$ & $2(d^2 + 2dr)$ & $2dk$ & $12dk$ & $2bsk$ \\
        \bottomrule
    \end{tabular}}}
    \label{tab:theoretical_memory}
    \vspace{-10pt}
\end{table}

\subsection{Multi-task Performance for Advanced Model Merging Techniques}

Extending the results in Section \ref{sec:multi-task}, we also evaluate two advanced model merging techniques, TIES-MERGING \cite{yadav2023ties} and DARE \cite{yu2024language}, for merging LoRA components from different domains. The results are listed in Tables \ref{tab:multi_tries_results} and \ref{tab:multi_dare_results}, respectively. Overall, both TIES-MERGING and DARE outperform naive weight averaging by resolving parameter conflicts through intelligent selection (TIES' sign consensus and trimming) and selective rescaling (DARE's dropout-based redundancy elimination). Similar to weight averaging, these results demonstrate that FlyLoRA consistently surpasses LoRA$_{(r=8)}$, LoRA$_{(r=32)}$, and Split-LoRA$_{(4 \times 8)}$ across all comparisons, highlighting the robustness of FlyLoRA's near-orthogonality property in reducing inter-task decoupling and further enhancing model merging performance.

We also include comparison with more advanced model merging techniques in Table \ref{tab:multi_knots_llora}. KnOTS \cite{stoica2024model} and L-LoRA \cite{tang2023parameter} are both built upon LoRA$_{(r=32)}$. The results suggest that they achieve comparable performance to FlyLoRA, with each method excelling on different datasets. It is noteworthy that FlyLoRA is not a competitor to these methods; rather, they can be used in a plug-and-play manner with FlyLoRA to further improve performance after merging.
\begin{table}[t]
    \centering
    \caption{
    \textbf{Multi-task Performance Comparison Before and After Parameter Merging Using TIES-MERGING.} We evaluate LoRA variants across MMLU, ScienceQA, GSM8K (accuracy), and HumanEval (Pass@k) benchmarks. The table shows performance before merging, after merging, and the relative performance drop ($\Delta$\%). The best results are highlighted in \textbf{bold}.
    }
    \vspace{+2pt}
    \renewcommand\arraystretch{1.0}
    \resizebox{1.0\textwidth}{!}{
    \setlength{\tabcolsep}{1mm}{
    \begin{tabular}{ccccccccc}
        \midrule
        \multirow{2}{*}{\textbf{Model}} & 
        \multirow{2}{*}{\textbf{Method}} & 
        \multirow{2}{*}{\textbf{Merge Status}} & 
        \multirow{2}{*}{\textbf{MMLU}} & 
        \multirow{2}{*}{\textbf{ScienceQA}} & 
        \multirow{2}{*}{\textbf{GSM8K}} & 
        \multicolumn{3}{c}{\textbf{HumanEval}} \\
        & & & & & & \textbf{Pass@1} & \textbf{Pass@5} & \textbf{Pass@10} \\
        \midrule
        \multirow{13}{*}{Llama-3.1-8B} 
            & \multirow{3}{*}{LoRA$_{(r=8)}$} 
            & Before &36.53\scriptsize{$\pm$0.40} &91.39\scriptsize{$\pm$0.55} &55.34\scriptsize{$\pm$0.24} &29.13\scriptsize{$\pm$0.56} &52.28\scriptsize{$\pm$1.24} &61.67\scriptsize{$\pm$0.61} \\
            & & After &31.83\scriptsize{$\pm$0.34} &35.96\scriptsize{$\pm$1.46} &27.53\scriptsize{$\pm$1.08} &18.45\scriptsize{$\pm$1.47} &45.52\scriptsize{$\pm$0.84} &56.63\scriptsize{$\pm$1.24} \\
            & & $\Delta$ (\%) &-4.70 &-55.43 &-27.81 &-10.68 &-6.76 &-5.04 \\
            \cmidrule{2-9}
            & \multirow{3}{*}{LoRA$_{(r=32)}$} 
            & Before &38.93\scriptsize{$\pm$1.04} &94.01\scriptsize{$\pm$0.17} &56.25\scriptsize{$\pm$0.29} &30.37\scriptsize{$\pm$1.06} &54.37\scriptsize{$\pm$0.39} &64.02\scriptsize{$\pm$0.94} \\
            & & After &34.45\scriptsize{$\pm$1.73} &36.63\scriptsize{$\pm$0.79} &26.59\scriptsize{$\pm$0.47} &20.98\scriptsize{$\pm$1.13} &47.15\scriptsize{$\pm$0.82} &59.97\scriptsize{$\pm$1.04} \\
            & & $\Delta$ (\%) &-4.48 &-57.38 &-29.66 &-9.39 &-7.22 &-4.05 \\
            \cmidrule{2-9}
            & \multirow{3}{*}{Split-LoRA$_{(4\times8)}$} 
            & Before &38.44\scriptsize{$\pm$0.69} &92.41\scriptsize{$\pm$0.54} &55.65\scriptsize{$\pm$0.47} &31.28\scriptsize{$\pm$1.52} &54.16\scriptsize{$\pm$1.12} &63.94\scriptsize{$\pm$0.89} \\
            & & After &33.92\scriptsize{$\pm$0.57} &40.81\scriptsize{$\pm$1.34} &29.02\scriptsize{$\pm$0.82} &22.02\scriptsize{$\pm$0.24} &46.32\scriptsize{$\pm$0.72} &59.72\scriptsize{$\pm$0.25} \\
            & & $\Delta$ (\%) &-4.52 &-51.60 &-26.63 &-9.26 &-7.84 &-4.22 \\
            \cmidrule{2-9}
            & \multirow{3}{*}{FlyLoRA$_{(k=8)}$} 
            & Before &40.88\scriptsize{$\pm$1.61} &94.15\scriptsize{$\pm$0.36} &58.76\scriptsize{$\pm$0.74} &36.88\scriptsize{$\pm$1.91} &62.40\scriptsize{$\pm$1.82} &73.34\scriptsize{$\pm$1.24} \\
            & & After &39.03\scriptsize{$\pm$1.31} &54.82\scriptsize{$\pm$0.46} &39.12\scriptsize{$\pm$1.02} &33.37\scriptsize{$\pm$0.62} &57.36\scriptsize{$\pm$1.41} &70.35\scriptsize{$\pm$0.39} \\
            & &\cellcolor{gray!20}\textbf{$\Delta$ (\%)} &\cellcolor{gray!20}\textbf{-1.85} &\cellcolor{gray!20}\textbf{-39.33} &\cellcolor{gray!20}\textbf{-19.64} &\cellcolor{gray!20}\textbf{-3.51} &\cellcolor{gray!20}\textbf{-5.04} &\cellcolor{gray!20}\textbf{-2.99} \\
        \midrule
        \multirow{13}{*}{Qwen-2.5-7B} 
            & \multirow{3}{*}{LoRA$_{(r=8)}$} 
            & Before &49.84\scriptsize{$\pm$0.56} &92.84\scriptsize{$\pm$0.13} &77.01\scriptsize{$\pm$0.32} &47.20\scriptsize{$\pm$1.54} &78.89\scriptsize{$\pm$0.36} &85.94\scriptsize{$\pm$0.64} \\
            & & After &44.98\scriptsize{$\pm$0.72} &61.69\scriptsize{$\pm$1.34} &81.89\scriptsize{$\pm$1.04} &23.37\scriptsize{$\pm$1.24} &68.96\scriptsize{$\pm$1.22} &81.25\scriptsize{$\pm$0.35} \\
            & & $\Delta$ (\%) &-4.86 &-31.15 &+4.88 &-23.83 &-9.93 &-4.69 \\
            \cmidrule{2-9}
            & \multirow{3}{*}{LoRA$_{(r=32)}$} 
            & Before &52.07\scriptsize{$\pm$0.31} &95.01\scriptsize{$\pm$0.21} &79.23\scriptsize{$\pm$0.22} &52.87\scriptsize{$\pm$1.79} &81.67\scriptsize{$\pm$1.14} &87.80\scriptsize{$\pm$0.72} \\
            & & After &35.92\scriptsize{$\pm$0.18} &58.02\scriptsize{$\pm$0.27} &83.98\scriptsize{$\pm$0.62} &24.79\scriptsize{$\pm$1.08} &67.50\scriptsize{$\pm$1.13} &80.35\scriptsize{$\pm$1.46} \\
            & & $\Delta$ (\%) &-16.15 &-36.99 &+4.75 &-28.08 &-14.17 &-7.45 \\
            \cmidrule{2-9}
            & \multirow{3}{*}{Split-LoRA$_{(4\times8)}$} 
            & Before &50.68\scriptsize{$\pm$1.06} &93.08\scriptsize{$\pm$0.41} &77.12\scriptsize{$\pm$0.76} &48.65\scriptsize{$\pm$1.18} &79.30\scriptsize{$\pm$0.91} &86.05\scriptsize{$\pm$0.44} \\
            & & After &44.96\scriptsize{$\pm$0.65} &60.59\scriptsize{$\pm$0.26} &81.82\scriptsize{$\pm$0.20} &23.53\scriptsize{$\pm$0.74} &67.59\scriptsize{$\pm$0.14} &82.34\scriptsize{$\pm$0.69} \\
            & & $\Delta$ (\%) &-5.72 &-32.49 &+4.70 &-25.12 &-11.71 &-3.71 \\
            \cmidrule{2-9}
            & \multirow{3}{*}{FlyLoRA$_{(k=8)}$} 
            & Before &53.68\scriptsize{$\pm$0.47} &95.55\scriptsize{$\pm$0.18} &80.82\scriptsize{$\pm$0.56} &54.34\scriptsize{$\pm$2.13} &82.85\scriptsize{$\pm$0.52} &89.63\scriptsize{$\pm$0.55} \\
            & & After &60.51\scriptsize{$\pm$0.37} &73.46\scriptsize{$\pm$0.80} &86.24\scriptsize{$\pm$0.31} &35.37\scriptsize{$\pm$0.47} &76.05\scriptsize{$\pm$1.25} &87.97\scriptsize{$\pm$1.09} \\
            & &\cellcolor{gray!20}\textbf{$\Delta$ (\%)} &\cellcolor{gray!20}\textbf{+6.83} &\cellcolor{gray!20}\textbf{-22.09} &\cellcolor{gray!20}\textbf{+5.42} &\cellcolor{gray!20}\textbf{-18.97} &\cellcolor{gray!20}\textbf{-6.80} &\cellcolor{gray!20}\textbf{-1.66} \\
        \midrule
    \end{tabular}}}
    \label{tab:multi_tries_results}
    \vspace{-10pt}
\end{table}
\begin{table}[t]
    \centering
    \caption{
    \textbf{Multi-task Performance Comparison Before and After Parameter Merging Using DARE.} We evaluate LoRA variants across MMLU, ScienceQA, GSM8K (accuracy), and HumanEval (Pass@k) benchmarks. The table shows performance before merging, after merging, and the relative performance drop ($\Delta$\%). The best results are highlighted in \textbf{bold}.
    }
    \vspace{+2pt}
    \renewcommand\arraystretch{1.0}
    \resizebox{1.0\textwidth}{!}{
    \setlength{\tabcolsep}{1mm}{
    \begin{tabular}{ccccccccc}
        \midrule
        \multirow{2}{*}{\textbf{Model}} & 
        \multirow{2}{*}{\textbf{Method}} & 
        \multirow{2}{*}{\textbf{Merge Status}} & 
        \multirow{2}{*}{\textbf{MMLU}} & 
        \multirow{2}{*}{\textbf{ScienceQA}} & 
        \multirow{2}{*}{\textbf{GSM8K}} & 
        \multicolumn{3}{c}{\textbf{HumanEval}} \\
        & & & & & & \textbf{Pass@1} & \textbf{Pass@5} & \textbf{Pass@10} \\
        \midrule
        \multirow{13}{*}{Llama-3.1-8B} 
            & \multirow{3}{*}{LoRA$_{(r=8)}$} 
            & Before &36.53\scriptsize{$\pm$0.40} &91.39\scriptsize{$\pm$0.55} &55.34\scriptsize{$\pm$0.24} &29.13\scriptsize{$\pm$0.56} &52.28\scriptsize{$\pm$1.24} &61.67\scriptsize{$\pm$0.61} \\
            & & After &31.24\scriptsize{$\pm$0.40} &34.37\scriptsize{$\pm$1.25} &26.76\scriptsize{$\pm$1.10} &17.35\scriptsize{$\pm$1.32} &45.49\scriptsize{$\pm$0.96} &57.24\scriptsize{$\pm$1.36} \\
            & & $\Delta$ (\%) &-5.29 &-57.02 &-28.58 &-11.78 &-6.79 &-4.43 \\
            \cmidrule{2-9}
            & \multirow{3}{*}{LoRA$_{(r=32)}$} 
            & Before &38.93\scriptsize{$\pm$1.04} &94.01\scriptsize{$\pm$0.17} &56.25\scriptsize{$\pm$0.29} &30.37\scriptsize{$\pm$1.06} &54.37\scriptsize{$\pm$0.39} &64.02\scriptsize{$\pm$0.94} \\
            & & After &34.75\scriptsize{$\pm$0.83} &37.56\scriptsize{$\pm$0.59} &26.89\scriptsize{$\pm$0.78} &19.36\scriptsize{$\pm$1.38} &46.67\scriptsize{$\pm$0.86} &59.85\scriptsize{$\pm$0.67} \\
            & & $\Delta$ (\%) &-4.18 &-56.45 &-29.36 &-11.01 &-7.70 &-4.17 \\
            \cmidrule{2-9}
            & \multirow{3}{*}{Split-LoRA$_{(4\times8)}$} 
            & Before &38.44\scriptsize{$\pm$0.69} &92.41\scriptsize{$\pm$0.54} &55.65\scriptsize{$\pm$0.47} &31.28\scriptsize{$\pm$1.52} &54.16\scriptsize{$\pm$1.12} &63.94\scriptsize{$\pm$0.89} \\
            & & After &34.02\scriptsize{$\pm$0.24} &38.58\scriptsize{$\pm$0.48} &28.63\scriptsize{$\pm$0.72} &23.52\scriptsize{$\pm$1.43} &46.84\scriptsize{$\pm$0.30} &60.30\scriptsize{$\pm$0.41} \\
            & & $\Delta$ (\%) &-4.42 &-53.83 &-27.02 &-7.76 &-7.32 &-3.64 \\
            \cmidrule{2-9}
            & \multirow{3}{*}{FlyLoRA$_{(k=8)}$} 
            & Before &40.88\scriptsize{$\pm$1.61} &94.15\scriptsize{$\pm$0.36} &58.76\scriptsize{$\pm$0.74} &36.88\scriptsize{$\pm$1.91} &62.40\scriptsize{$\pm$1.82} &73.34\scriptsize{$\pm$1.24} \\
            & & After &39.37\scriptsize{$\pm$1.02} &52.34\scriptsize{$\pm$0.35} &38.20\scriptsize{$\pm$1.52} &33.34\scriptsize{$\pm$0.83} &57.14\scriptsize{$\pm$1.37} &70.24\scriptsize{$\pm$0.42} \\
            & &\cellcolor{gray!20}\textbf{$\Delta$ (\%)} &\cellcolor{gray!20}\textbf{-1.51} &\cellcolor{gray!20}\textbf{-41.81} &\cellcolor{gray!20}\textbf{-20.56} &\cellcolor{gray!20}\textbf{-3.54} &\cellcolor{gray!20}\textbf{-5.26} &\cellcolor{gray!20}\textbf{-3.10} \\
        \midrule
        \multirow{13}{*}{Qwen-2.5-7B} 
            & \multirow{3}{*}{LoRA$_{(r=8)}$} 
            & Before &49.84\scriptsize{$\pm$0.56} &92.84\scriptsize{$\pm$0.13} &77.01\scriptsize{$\pm$0.32} &47.20\scriptsize{$\pm$1.54} &78.89\scriptsize{$\pm$0.36} &85.94\scriptsize{$\pm$0.64} \\
            & & After &45.20\scriptsize{$\pm$0.40} &61.39\scriptsize{$\pm$1.32} &81.98\scriptsize{$\pm$0.86} &23.49\scriptsize{$\pm$1.02} &69.04\scriptsize{$\pm$0.17} &81.22\scriptsize{$\pm$0.27} \\
            & & $\Delta$ (\%) &-4.64 &-31.45 &+4.97 &-23.71 &-9.85 &-4.72 \\
            \cmidrule{2-9}
            & \multirow{3}{*}{LoRA$_{(r=32)}$} 
            & Before &52.07\scriptsize{$\pm$0.31} &95.01\scriptsize{$\pm$0.21} &79.23\scriptsize{$\pm$0.22} &52.87\scriptsize{$\pm$1.79} &81.67\scriptsize{$\pm$1.14} &87.80\scriptsize{$\pm$0.72} \\
            & & After &35.27\scriptsize{$\pm$1.68} &56.79\scriptsize{$\pm$1.34} &84.07\scriptsize{$\pm$0.16} &24.35\scriptsize{$\pm$1.07} &67.74\scriptsize{$\pm$0.70} &80.12\scriptsize{$\pm$0.32} \\
            & & $\Delta$ (\%) &-16.80 &-38.22 &+4.80 &-28.52 &-13.93 &-7.68 \\
            \cmidrule{2-9}
            & \multirow{3}{*}{Split-LoRA$_{(4\times8)}$} 
            & Before &50.68\scriptsize{$\pm$1.06} &93.08\scriptsize{$\pm$0.41} &77.12\scriptsize{$\pm$0.76} &48.65\scriptsize{$\pm$1.18} &79.30\scriptsize{$\pm$0.91} &86.05\scriptsize{$\pm$0.44} \\
            & & After &43.56\scriptsize{$\pm$0.84} &62.15\scriptsize{$\pm$0.26} &81.82\scriptsize{$\pm$0.20} &23.79\scriptsize{$\pm$0.52} &68.74\scriptsize{$\pm$0.96} &81.53\scriptsize{$\pm$1.07} \\
            & & $\Delta$ (\%) &-7.12 &-30.93 &+4.70 &-24.86 &-10.56 &-4.52 \\
            \cmidrule{2-9}
            & \multirow{3}{*}{FlyLoRA$_{(k=8)}$} 
            & Before &53.68\scriptsize{$\pm$0.47} &95.55\scriptsize{$\pm$0.18} &80.82\scriptsize{$\pm$0.56} &54.34\scriptsize{$\pm$2.13} &82.85\scriptsize{$\pm$0.52} &89.63\scriptsize{$\pm$0.55} \\
            & & After &61.35\scriptsize{$\pm$0.47} &72.94\scriptsize{$\pm$0.21} &86.34\scriptsize{$\pm$0.36} &34.47\scriptsize{$\pm$0.44} &75.97\scriptsize{$\pm$0.70} &87.64\scriptsize{$\pm$1.04} \\
            & &\cellcolor{gray!20}\textbf{$\Delta$ (\%)} &\cellcolor{gray!20}\textbf{+7.67} &\cellcolor{gray!20}\textbf{-22.61} &\cellcolor{gray!20}\textbf{+5.52} &\cellcolor{gray!20}\textbf{-19.87} &\cellcolor{gray!20}\textbf{-6.88} &\cellcolor{gray!20}\textbf{-1.99} \\
        \midrule
    \end{tabular}}}
    \label{tab:multi_dare_results}
    \vspace{-10pt}
\end{table}
\begin{table}[h]
    \centering
    \caption{
        \textbf{Multi-task Performance Comparison with More Advanced Merging Techniques using Qwen-2.5-7B.} We evaluate different methods across MMLU, ScienceQA, GSM8K (accuracy), and HumanEval (Pass@k) benchmarks. The table shows the relative performance drop ($\Delta$\%) before and after merging. The best results are highlighted in \textbf{bold}.
    }
    \vspace{+2pt}
    \resizebox{0.6\textwidth}{!}{
    \setlength{\tabcolsep}{1mm}{
    \begin{tabular}{ccccc}
        \toprule
        \textbf{Method} & \textbf{MMLU} & \textbf{ScienceQA} & \textbf{GSM8K} & \textbf{HumanEval} \\
        \midrule
        FlyLoRA & +6.55 & -23.77 & +4.80 & -21.23 \\
        KnOTS & +10.76 & -26.85 & +4.68 & -23.37 \\
        L-LoRA & +4.51 & -22.48 & +4.74 & -20.85 \\
        KnOTS+FlyLoRA & \textbf{+11.47} & -23.41 & \textbf{+5.25} & -20.69 \\
        L-LoRA+FlyLoRA & +7.65 & \textbf{-21.42} & +5.02 & \textbf{-19.85} \\
        \bottomrule
    \end{tabular}}}
    \label{tab:multi_knots_llora}
    \vspace{-10pt}
\end{table}

\subsection{Additional Ablation Studies on Load-Balancing Strategies}
\begin{table}[h]
    \centering
    \caption{
        \textbf{Performance Comparison of Different Load-Balancing Strategies.} Evaluation on MMLU benchmark using Llama-3.1-8B.
    }
    \vspace{+2pt}
    \resizebox{0.4\textwidth}{!}{
    \setlength{\tabcolsep}{1mm}{
    \begin{tabular}{cc}
        \toprule
        \textbf{Load-Balancing Strategy} & \textbf{Accuracy (\%)} \\
        \midrule
        Loss-Free & \textbf{40.88\scriptsize{$\pm$1.61}} \\
        Loss-Controlled & 40.59\scriptsize{$\pm$0.51} \\
        No Load-Balancing & 37.56\scriptsize{$\pm$2.87} \\
        \bottomrule
    \end{tabular}}}
    \label{tab:router_ablation}
    \vspace{-10pt}
\end{table}
We compare experimental results using different load-balancing strategies. In Section \ref{sec:formulation}, we employ an easy-to-implement loss-free balancing strategy. This loss-agnostic approach effectively achieves load balancing with negligible computational overhead and memory footprint. Simultaneously, other loss-controlled load-balancing strategies like \cite{fedus2022switch} are widely used in MoE-like structures. A comparison of them is shown in Table \ref{tab:router_ablation}. Our results show that different routing strategies achieve similar effects in performance. While FlyLoRA requires load-balancing strategies, it is not sensitive to specific methods.

\subsection{Additional Ablation Studies on K-Selection Strategies}
To evaluate the impact of activation selection, we analyze different K-selection approaches in our experiments. In neuroscience, the fly olfactory circuit implements a ``winner-take-all'' strategy, simulated through top-$k$ selection based on activation values across dimensions. To validate top-$k$'s effectiveness, we test random-$k$ selection and full activation (without selection) as baselines. Results in Table \ref{tab:k_selection_ablation} show that both top-$k$ and random-$k$ outperform full activation, confirming sparse activation mitigates intra-task interference. Crucially, top-$k$ surpasses random-$k$ because random selection cannot prioritize the most informative dimensions, leading to suboptimal performance. These findings collectively demonstrate that biologically inspired top-$k$ activation optimally balances efficiency and task-specific feature selection.
\begin{table}[h]
    \centering
    \caption{
        \textbf{Performance Comparison of Different K-Selection Strategies.} Evaluation on MMLU benchmark using Llama-3.1-8B.
    }
    \vspace{+2pt}
    \resizebox{0.4\textwidth}{!}{
    \setlength{\tabcolsep}{1mm}{
    \begin{tabular}{cc}
        \toprule
        \textbf{K-Selection Strategy} & \textbf{Accuracy (\%)} \\
        \midrule
        top-$k$ & \textbf{40.88\scriptsize{$\pm$1.61}} \\
        random-$k$ & 40.02\scriptsize{$\pm$0.26} \\
        full activation & 39.40\scriptsize{$\pm$1.14} \\
        \bottomrule
    \end{tabular}}}
    \label{tab:k_selection_ablation}
    \vspace{-10pt}
\end{table}

\subsection{Additional Ablation Studies on Matrix A initialization Schemes}
We further compare three methods for generating the sparse projection $\bm{A}$ with $\frac{p}{r}$ sparsity:
\begin{enumerate}[topsep=0pt,itemsep=0ex,leftmargin=3ex]
    \item Gaussian (our default): Each non-zero entry is drawn from $\mathcal{N}(0, \frac{1}{r^2})$.
    \item Rademacher (non-Gaussian): Each non-zero entry is $\pm\frac{1}{r}$ with equal probability.
    \item FJLT \cite{ailon2009fast} (structured projection): $\bm{A}=\bm{PHD}$, where $\bm{D}$ is a random diagonal matrix with independent Rademacher variables on its diagonal, $\bm{H}$ is a normalized Hadamard matrix, and $\bm{P}$ enforces the $\frac{p}{r}$ sparsity.
    \item Two-Phase (briefly-learned): The non-zero entries of $\bm{A}$ are trainable for 5\% of total steps as warm-up, then frozen for the remainder.
\end{enumerate}
The results, shown in Table \ref{tab:A_initialization_ablation}, indicate that almost all variants perform similarly. Non-Gaussian, structured, or briefly-learned initializations have little impact, except that the briefly-learned scheme shows a noticeable drop after merging. This demonstrates that FlyLoRA is robust to the choice of initialization scheme for matrix $\bm{A}$, and that learning may break the approximate orthogonality of the random matrix, making it unsuitable.
\begin{table}[t]
    \centering
    \caption{
        \textbf{Performance Comparison of Different A Initialization Strategies.} Single-task and multi-task accuracy comparison on MMLU using Llama-3.1-8B.
    }
    \vspace{+2pt}
    \resizebox{0.5\textwidth}{!}{
    \setlength{\tabcolsep}{1mm}{
    \begin{tabular}{ccc}
        \toprule
        \textbf{A Initialization Strategy} & \textbf{Before} & $\bm\Delta$\textbf{after merging} \\
        \midrule
        Gaussian & 40.88\scriptsize{$\pm$1.61} & -2.02 \\
        Rademacher & 40.42\scriptsize{$\pm$0.23} & -2.35 \\
        FJLT & 40.57\scriptsize{$\pm$1.34} & -2.50 \\
        Two-Phase & 40.76\scriptsize{$\pm$1.04} & -4.86 \\
        \bottomrule
    \end{tabular}}}
    \label{tab:A_initialization_ablation}
    \vspace{-10pt}
\end{table}

\subsection{Analysis of the Performance Gap Between Merged and Non-merged Scenarios}
In Tables \ref{tab:multi_results}, \ref{tab:multi_tries_results}, and \ref{tab:multi_dare_results}, we can see that ScienceQA usually demonstrates a large performance drop after merging for all LoRA variants. Intuitively, the four tasks---general knowledge understanding (MMLU), scientific question answering (ScienceQA), mathematical reasoning (GSM8K), and code generation (HumanEval)---represent significantly different distributions, so merging their adapters is prone to substantial conflicts.

Empirically, following \cite{stoica2024model}, we use centered kernel alignment (CKA) \cite{kornblith2019similarity} to quantify the alignment between the output representations of each single-task adapter and the merged adapter. A higher CKA indicates better output alignment, which is likely the inherent reason, and therefore, results in a smaller accuracy drop after merging. Table \ref{tab:cka} reports both CKA and accuracy drop ($\Delta$) on Llama-3.1-8B. Since there is no apparent difference between LoRA$_{(r=8)}$, LoRA$_{(r=32)}$, and Split-LoRA$_{(4\times8)}$ in model merging, we use LoRA$_{(r=8)}$ as a representative to compare with FlyLoRA$_{(k=8)}$. We observe that tasks with lower CKA (especially ScienceQA and GSM8K) suffer the largest accuracy drops. FlyLoRA consistently yields higher CKA than LoRA, which aligns with its consistently smaller $\Delta$. This micro-level analysis corroborates why FlyLoRA outperforms LoRA in heterogeneous-task merging.
\begin{table}[t]
    \centering
    \caption{
        \textbf{CKA and Corresponding Accuracy Drop ($\Delta$) Between Single-Task Adapter and Merged Model.} Evaluating using Llama-3.1-8B.
    }
    \vspace{+2pt}
    \resizebox{0.6\textwidth}{!}{
    \setlength{\tabcolsep}{1mm}{
    \begin{tabular}{cccccc}
        \toprule
        \textbf{Method} & \textbf{Task} & \textbf{MMLU} & \textbf{ScienceQA} & \textbf{GSM8K} & \textbf{HumanEval} \\
        \midrule
        \multirow{2}{*}{LoRA$_{(r=8)}$} 
            & CKA & 0.78 & 0.39 & 0.58 & 0.75 \\
            & $\Delta$ & -6.48 & -60.34 & -30.15 & -13.04 \\
        \midrule
        \multirow{2}{*}{FlyLoRA$_{(k=8)}$} 
            & CKA & \textbf{0.85} & \textbf{0.53} & \textbf{0.71} & \textbf{0.84} \\
            & $\Delta$ & \textbf{-2.02} & \textbf{-43.05} & \textbf{-21.81} & \textbf{-4.27} \\
        \bottomrule
    \end{tabular}}}
    \label{tab:cka}
    \vspace{-10pt}
\end{table}

\section{Detailed Experimental Setting} \label{app:experimental_setting}

\subsection{Datasets}
To comprehensively evaluate the effectiveness of our proposed method across diverse domains and task types, we conduct extensive experiments on five carefully selected benchmarks. These datasets span critical capabilities including general knowledge reasoning, scientific understanding, mathematical problem solving, and code generation. Table \ref{tab:dataset} summarizes the key characteristics of each dataset, while detailed descriptions are provided below:
\begin{itemize}
    \item \textbf{MMLU}~\cite{hendryckstest2021} serves as a comprehensive benchmark for evaluating broad knowledge understanding and reasoning capabilities. It comprises multiple-choice questions spanning 57 distinct academic subjects, ranging from elementary mathematics and US history to computer science and professional law. The diversity of domains makes it particularly suitable for assessing model generalization across different knowledge types.

    \item \textbf{ScienceQA}~\cite{lu2022learn} provides a multimodal framework for science question answering, with content derived from elementary and high school curricula aligned with California Common Core Content Standards. The questions originate from IXL Learning's expert-curated educational resources. Following the established practice in \cite{li2024mixlora}, we utilize only the textual components to focus on linguistic understanding.

    \item \textbf{GSM8K}~\cite{cobbe2021gsm8k} offers 8,500 high-quality grade school mathematics word problems that demand multi-step arithmetic reasoning. Each problem is accompanied by a detailed, step-by-step solution, making it ideal for evaluating logical reasoning and procedural accuracy in mathematical contexts.

    \item \textbf{CodeAlpaca-20k}~\cite{codealpaca} contains 20,022 synthetically generated instruction-response pairs specifically designed for code-related tasks. This dataset facilitates effective instruction tuning for programming applications by providing diverse coding prompts paired with corresponding solutions.

    \item \textbf{HumanEval}~\cite{chen2021evaluating} consists of 164 hand-crafted Python programming problems developed to assess functional correctness in code generation. Crucially, these problems were manually created to prevent data contamination, ensuring they do not appear in the training corpora of existing code generation models.
\end{itemize}
\begin{table}[h]
    \centering
    \caption{
        \textbf{Details of MMLU, ScienceQA, GSM8K, CodeAlpaca and HumanEval Datasets.} We list the number of training and testing samples and task types for the following datasets used in our experiments.
    }
    \vspace{+2pt}
    \resizebox{0.75\textwidth}{!}{
    \setlength{\tabcolsep}{1mm}{
    \begin{tabular}{cccc}
        \toprule
        \textbf{Dataset} & \textbf{Training Samples} & \textbf{Testing Samples} & \textbf{Task Types} \\
        \midrule
        MMLU \cite{hendryckstest2021} & 99,842 & 14,042 & Multiple Choice \\
        ScienceQA \cite{lu2022learn} & 12,726 & 4,241 & Multiple Choice \\
        GSM8K \cite{cobbe2021gsm8k} & 7,473 & 1,319 & Math Problems \\
        CodeAlpaca-20k \cite{codealpaca} & 20,022 & $-$ & Code Instruction \\
        HumanEval \cite{chen2021evaluating} & $-$ & 164 & Code Generation \\
        \bottomrule
    \end{tabular}}}
    \label{tab:dataset}
    \vspace{-10pt}
\end{table}

\subsection{Training Configuration}
This section elaborates on the experimental setup and hyperparameter configurations employed throughout our study. Table \ref{tab:general_config} documents the shared training parameters applied consistently across all backbone models and datasets. To address the specific requirements of different tasks and model architectures, we additionally provide dataset-specific and model-specific configurations in Table \ref{tab:specific_config}, including learning rate schedules, batch size adjustments, and task-specific optimization strategies.
\begin{table}[t]
    \centering
    \caption{\textbf{General Training Hyperparameters for FlyLoRA.} Shared configuration across all experiments, including rank settings, optimizer details, and architectural choices.}
    \label{tab:general_config}
    \vspace{+2pt}
    \resizebox{0.7\textwidth}{!}{
    \begin{tabular}{lc}
        \toprule
        \textbf{Parameter} & \textbf{Value} \\
        \midrule
        Total rank ($r$) & 32 \\
        Scaling factor ($\alpha$) & 64 \\
        Activated rank & 8 \\
        Target modules & \texttt{\{q,k,v,o,gate,down,up\}\_proj} \\
        Optimizer & AdamW \\
        Warmup ratio & 0.01 \\
        Gradient accumulated batch & 128 \\
        Dropout rate & 0.00 \\
        \bottomrule
    \end{tabular}}
    \vspace{-10pt}
\end{table}
\begin{table}[t]
    \centering
    \caption{\textbf{Dataset-Specific and Model-Specific Training Configurations for FlyLoRA.} Task-optimized settings for Llama-3.1-8B and Qwen-2.5-7B across four benchmarks, showing variations in epoch counts, learning rates, and sequence lengths based on dataset characteristics and model requirements.}
    \label{tab:specific_config}
    \vspace{+2pt}
    \resizebox{0.9\textwidth}{!}{
    \begin{tabular}{clcccc}
        \toprule
        \textbf{Model} & \textbf{Parameter} & \textbf{MMLU} & \textbf{ScienceQA} & \textbf{GSM8K} & \textbf{CodeAlpaca} \\
        \midrule
        \multirow{4}{*}{Llama-3.1-8B}
        & Epochs & 1 & 20 & 1 & 2 \\
        & Learning rate & $3\times10^{-4}$ & $3\times10^{-4}$ & $3\times10^{-4}$ & $3\times10^{-4}$ \\
        & Max sequence length & 128 & 256 & 512 & 512 \\
        & micro batch size & 8 & 8 & 8 & 8 \\
        \midrule
        \multirow{4}{*}{Qwen-2.5-7B}
        & Epochs & 1 & 20 & 1 & 2 \\
        & Learning rate & $3\times10^{-4}$ & $3\times10^{-4}$ & $3\times10^{-4}$ & $6\times10^{-4}$ \\
        & Max sequence length & 128 & 256 & 512 & 512 \\
        & micro batch size & 8 & 8 & 8 & 8 \\
        \bottomrule
    \end{tabular}}
    \vspace{-10pt}
\end{table}

\subsection{Split-LoRA} \label{App:Split-LoRA}
We implement Split-LoRA as a representative MoE-based LoRA method, following the general framework described in Section \ref{Sec:MoELoRA}. In our experiments, we incorporate a sigmoid activation function in the router to normalize expert selection scores. Thus, the gating function operates as $\bm{G}(\bm{x})=\text{sigmoid}(\text{top-}k(\bm{W}_g\bm{x}))$, which ensures differentiable routing while maintaining the sparsity of expert activation. This configuration allows Split-LoRA to serve as a representative baseline for evaluating the effectiveness of MoE structures in LoRA.

\subsection{Environments}
Most experiments were conducted on a Linux server running Ubuntu 20.04.4 LTS, equipped with an Intel(R) Xeon(R) Platinum 8358P CPU at 2.60GHz and 8 NVIDIA GeForce RTX 3090 GPUs, using CUDA version 11.7. Experiments with Qwen-2.5-14B were conducted on a machine with 8 NVIDIA A100 GPUs.

\section{Limitations and Future Work} \label{App:limit_future}
In FlyLoRA, matrix $\bm{A}$ is randomly initialized and frozen during training, but there may still be room for improvement. Recent neuroscience studies \cite{dasgupta2017neural} suggest that $\bm{A}$ need not be entirely frozen and random, indicating potential for more bio-inspired mechanisms to enhance task decoupling through an adaptable version of $\bm{A}$. Moreover, recent works suggest that component-wise interpretability \cite{zhu2025patchwise} and spectral modulation \cite{sun2024unleashing} could inspire adaptive or frequency-aware modifications of $\bm{A}$ in FlyLoRA to improve efficiency, robustness, and task decoupling.

Recently, RL fine-tuning for LLMs has emerged as a promising approach that significantly enhances their reasoning ability \cite{guo2025deepseek}. However, stabilizing MoE RL training remains an open question \cite{zheng2025group}, and further exploration will focus on the integration of FlyLoRA with RL training \cite{qu2025can} and potentially extending it to offline policy optimization \cite{mao2023supporteda, mao2023supportedb, mao2024offline, mao2024doubly, qu2023hokoff, shao2023counterfactual}. Additionally, integrating active data selection methods could be a promising direction to further improve data efficiency \cite{qu2025fast, wang2024greats, wang2025model, zou2025utility}.

\section{Broader Impact} \label{App:broader_impact}
Our proposed FlyLoRA resolves the trade-off between parameter interference and efficiency in MoE-based LoRA approaches. Additionally, this efficient decoupling mechanism, which is inspired by fly olfactory circuits, can be applied across various domains, helping researchers and developers leverage more powerful LoRA fine-tuning strategies. On the other hand, FlyLoRA could potentially be misused to fine-tune LLMs that exhibit biases or generate harmful content. We recommend implementing model access controls and bias-monitoring frameworks when deploying this technique.

\end{document}